\documentclass[conference,final,]{IEEEtran}
\ifCLASSINFOpdf
\else
\fi

\usepackage{longtable,booktabs}
\usepackage{graphicx}
\makeatletter
\def\maxwidth{\ifdim\Gin@nat@width>\linewidth\linewidth
\else\Gin@nat@width\fi}
\makeatother
\let\Oldincludegraphics\includegraphics
\renewcommand{\includegraphics}[1]{\Oldincludegraphics[width=\maxwidth]{#1}}

\usepackage[unicode=true]{hyperref}

\hypersetup{
            pdftitle={Explainable Artificial Intelligence Based Fault Diagnosis and Insight Harvesting for Steel Plates Manufacturing},
            pdfkeywords={Automated Manufacturing, Predictive Maintenance, Explainable Artificial Intelligence, Insight Harvesting, Fault Diagnosis},
            pdfborder={0 0 0},
            breaklinks=true}
\usepackage{amsfonts}
\usepackage{booktabs}
\usepackage{longtable}
\usepackage{array}
\usepackage{multirow}
\usepackage{wrapfig}
\usepackage{float}
\usepackage{colortbl}
\usepackage{pdflscape}
\usepackage{tabu}
\usepackage{threeparttable}
\usepackage{threeparttablex}
\usepackage[normalem]{ulem}
\usepackage{makecell}
\usepackage{xcolor}

\begin{document}
%
\title{Explainable Artificial Intelligence Based Fault Diagnosis and Insight Harvesting for Steel Plates Manufacturing}


\author{

\IEEEauthorblockN{
Athar Kharal
}

Department of Mathematics\\
The Air University,
Pakistan
\\ atharkharal@gmail.com
}


%


\maketitle

\begin{abstract}
With the advent of Industry 4.0, Data Science and Explainable Artificial Intelligence (XAI) has received considerable intrest in recent literature. However, the entry threshold into XAI, in terms of computer coding and the requisite mathematical apparatus, is really high. For fault diagnosis of steel plates, this work reports on a methodology of incorporating XAI based insights into the Data Science process of development of high precision classifier. Using Synthetic Minority Oversampling Technique (SMOTE) and notion of medoids, insights from XAI tools viz.~Ceteris Peribus profiles, Partial Dependence and Breakdown profiles have been harvested. Additionally, insights in the form of IF-THEN rules have also been extracted from an optimized Random Forest and Association Rule Mining. Incorporating all the insights into a single ensamble classifier, a 10 fold cross validated performance of 94\% has been achieved. In sum total, this work makes three main contributions viz.: methodogly based upon utilization of medoids and SMOTE, of gleaning insights and incorporating into model development process. Secondly the insights themselves are contribution, as they benefit the human experts of steel manufacturing industry, and thirdly a high precision fault diagnosis classifier has been developed.\\
\end{abstract}



\maketitle


%
\IEEEpeerreviewmaketitle

\hypertarget{introduction}{%
\section{Introduction}\label{introduction}}

Data Science carries two senses of meaning namely `Science \emph{of} Data' and `Science \emph{by} Data'. In the sense of Science \emph{of} Data, new methods, techniques and insights for handling, manipulating and predictive modeling of data are developed and extended. In the second sense i.e.~`Science \emph{by} Data', data of a specific scientific domain is examined and knowledge is extracted from it for the betterment and enhancement of the original scientific domain. It is this second sense namely doing `Science \emph{by} Data', that this paper owes the spirit and motivation.

Insight, the structural understanding of phenomena, is the essence of human intelligence. Usualy, insight is distilled from the fusion of information from various sources e.g.~past experiences, analysis techniques and methods (Roscher et al. 2020). There is no reason to not to expect insights from Artificial Intelligence (AI) as well, which more appropriately may (in our view, must) be called Machine Intelligence (MI). AI has shown great strides in recent times. Passing through the stages of \emph{Symbolic} AI, \emph{Computational} AI, it has now reached to another milestone namely, \emph{Explainable} AI (XAI) or more appropriately Explainable Machine Learning (XML). Explainability has remained a much sought after feature of AI systems right from the era of Expert Systems hype of AI, mainly because even the best human experts needed assistance/explanation from the machine to reach the right decisions. Non-availability of explainability has also remained one of the major barriers in a wide-scale practical adoption of AI. Improved understanding of the inner working of a phenomenon and its machine representation i.e.~ML model leads to the correction of its weaknesses and flaws. Understanding of one subsytem, obtained from XAI, may also be transferred to another system thus `democratizing' the invaluable expertise in a cummulative manner. Black box models, though considerably successful in prediction tasks, but at the same time they present a totally opaque face when it comes to explanation and insight. In recent time the confluence of Data Science (as Science \emph{of} Data) with AI techniques has achieved great deal of success to peep into even a high precision predictive models of black-box genre. Clearly, it is an area located at the crossroads of data-fusion and information-fusion.

The understanding, structural relations and the relational insight achieved by data is just as good as the data itself. It cannot guarantee the causal scientific relationship of the external world. Nevertheless, an explainable AI model offers a viable hypothesis to begin with (cf.~Section 3 of (Roscher et al. 2020)). The journey of scientific enquiry, at least may be begun in most of the cases in the right direction. Furthermore, explainability also ensures that only meaningful variables are allowed to take part in final modelling tasks thus apporaching the ideal of a truly parsimonious model as well. Christina et. al.~(Christina B. Azodi, Jiliang Tang, and Shin-Han Shiu 2020) have highlighted the same point more aptly: \emph{``ML interpretation strategies mostly do not identify causal relationships between input features and labels. Instead, interpretations should be used to generate hypotheses of cause-effect relations that can be tested experimentally.''}

Industrial plant monitoring focuses upon optimization of resources and thus attempts to achieve minimal error costs, improved quality of production and safety of workplace. A timely, precise and indicative of the root-cause detections of defects or faults is fundamental to such monitoring. Compliance management also neciassitates the explainablity of the machine learning models. That is why due to heavy cost of production, in general all industry, and in particular steel production industry has shown great intrest and effort to reduce the hugely expensive faults of production. Fault diagnostics in conjuction with fault isolation and root-cause analysis also contributes to precautionary maintenance or the prognostics of the production system (Soylemezoglu, Jagannathan, and Saygin 2011). Due to cost of faults and the ushering of Industry 4.0, Internet of Things (IoT) and Cyber Physical System (CPS), interest in data analytics based solutions has increased to a remarkable degree.

Fault diagnosis within an industrial production setup is further important on two more counts. Obviously, one for maintaing the quality of production. This requires fast and precise, either human experts (Gabriel Schwartz and Ko Nishino 2020) or preferably intelligent digital solutions like machine and statistical learning based predictive models. As the human expertise becomes more and more scarce and costly, the automated digital solutions become even more relevant and in-demand. Second and more important utility of fault diagnosis is its use to identify the exact causes, variables and factors of production which are responsible for a particular fault. This usage of fault diagnosis, in fact, demands developing the insight into the production system through its various variables. It is exactly here that the present work is located. It aims to benefit the quality production of steel plates at industrial scale by enhancing the human insight of the fault producing factors and causes.

Steel plates fault diangosis has been studied by various researchers. In (Nkonyana et al. 2019), authors employed various models viz.~ANN (Kharal and Saleem 2012), SVM and Random Forests. Random Forest achieved the highest accuracy of 77.80\% while SVM followed by 73.60\%. So far the highest accuracy of prediction has been reported by Tian in (Tian, Fu, and Wu 2015). This SVM implementation reported a performance of 80.7443\%. However, authors also noted that other sample balancing techniques may bring better result than the one used. In the settings of smart cities, the fault reporting and diagnostics using linked data has also been studied in (Consoli et al. 2017). Authors of (E.L. Russell, L.H.Chiang, and R.D.Braatz 2000) have applied the Canonical Variate Statistic, residual space PCA, DPCA, and CVA to the Tennessee Eastman process simulator of a wide variety of faults occurring in a chemical plant. Kazemi et.al. (Kazemi, Hajian, and Kiani 2018) calculate the positions (ranking) of each classifier compared to the five datasets and calculate the average ranking, and showed the quality of the proposed meta-classifier to be the best.

This paper is organized as follows: It comprises two main parts. Section \ref{MLdetails} describes various details of machine learning experiments e.g.~Class balancing, optimization and learner comparisons. Section \ref{insights} focuses upon extraction of insights from the trained machine learning model. It extracts insights from XAI tools, namely Feature Importance (\ref{featimp}), Ceteris Peribus Profiles (\ref{cp}), Partial Dependence Profiles (\ref{pdp}) and Breakdown Profiles (\ref{bd}). Section \ref{rf} and \ref{arm} present the insights extracted from optimized random forest and association rules mining, respectively. Section \ref{mdl} describes how all the obtained insights may be incorporated into an ensamble learner. Finally Section \ref{discuss} presents the discussion and conclusions. A graphical summary is presented in the following flow diagram:

\includegraphics{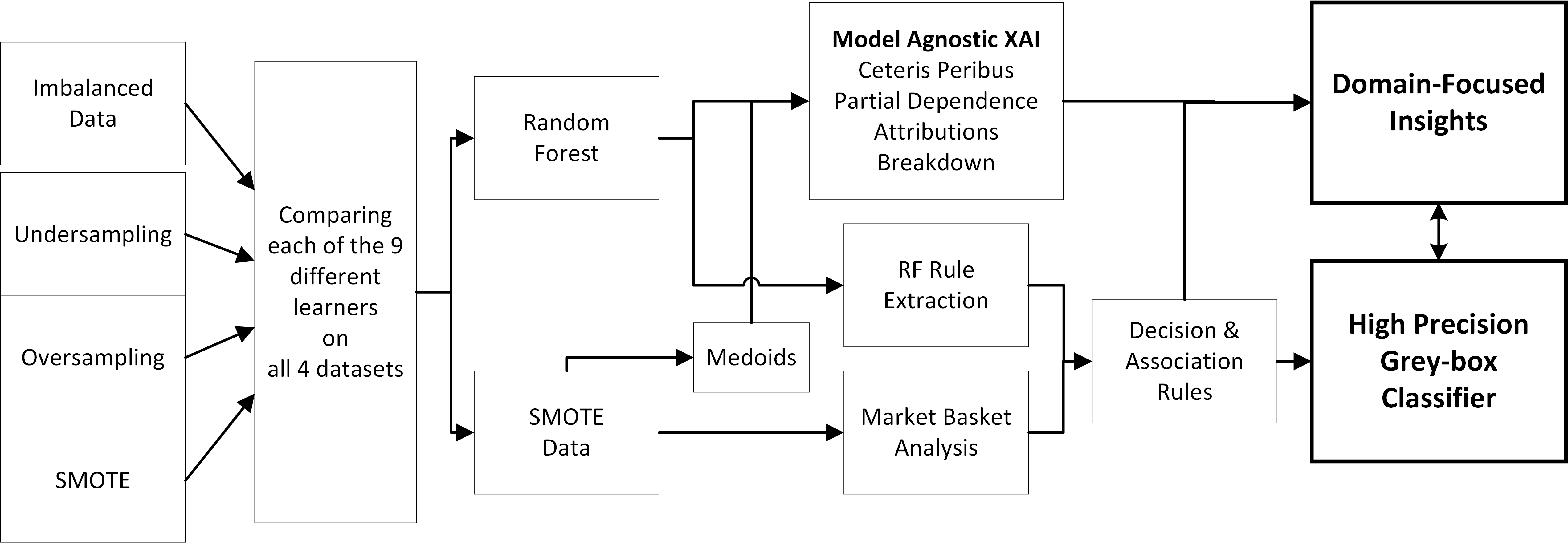}
\emph{Graphical Summary: Balancing of data has been followed by experimenting with 9 different classifier algorithms thus resulting into 36 different machine learning experiments. Optimized best performer, XAI and Association Rule Mining has been employed to develop an ensamble with high accuracy.}

\hypertarget{MLdetails}{%
\section{Machine Learning Details}\label{MLdetails}}

To correctly classify the type of surface defects of steel plates during industrial production, Semeion of Italy collected a data of 1941 faulty steel plates and recorded 27 observable characteristics/features of each of the faulty plate. This data is publically available through UCI (Semeion, n.d.). Data has 34 columns of which 27 are the observable variables/features and 7 binary columns each showing presence or absence of each of the seven types of faults. The last seven columns are one hot encoded classes, i.e.~if the plate fault is classified as (say) ``Stains'' there will be a 1 in that column and 0's in the other columns. However for the machine learning experiment requirements, following two changes have been made in present work:\\
1. Merged two factor variables viz.~`TypeOfSteel\_A300' and'TypeOfSteel\_A400' as a single variable named `TypeOfSteel',\\
2. Merged the 7 fault indicator variables into one variable showing all the 7 faults. This new target variable has been named as `Fault'.\\
Variable names of the dataset after the abovementioned two changes are given in Table \ref{tab:vars}:

\begin{table}[H]

\caption{\label{tab:vars}Variables of the Dataset}
\centering
\begin{tabular}[t]{ll}
\toprule
 & \\
\midrule
Edges\_Index & Outside\_X\_Index\\
Edges\_X\_Index & Pixels\_Areas\\
Edges\_Y\_Index & SigmoidOfAreas\\
Empty\_Index & Square\_Index\\
Fault & Steel\_Plate\_Thickness\\
\addlinespace
Length\_of\_Conveyer & Sum\_of\_Luminosity\\
Log\_X\_Index & TypeOfSteel\\
Log\_Y\_Index & X\_Maximum\\
LogOfAreas & X\_Minimum\\
Luminosity\_Index & X\_Perimeter\\
\addlinespace
Maximum\_of\_Luminosity & Y\_Maximum\\
Minimum\_of\_Luminosity & Y\_Minimum\\
Orientation\_Index & Y\_Perimeter\\
Outside\_Global\_Index & \\
\bottomrule
\end{tabular}
\end{table}

All the variables in dataset are numeric except `TypeOfSteel' and `Fault' which are categorical. `TypeOfSteel' indicates type to be either A300 or A400. `Fault' is the target variable containing 7 labels of fault names (cf.~Figure \ref{fig:imbalance}).

\begin{figure}[h]
\includegraphics{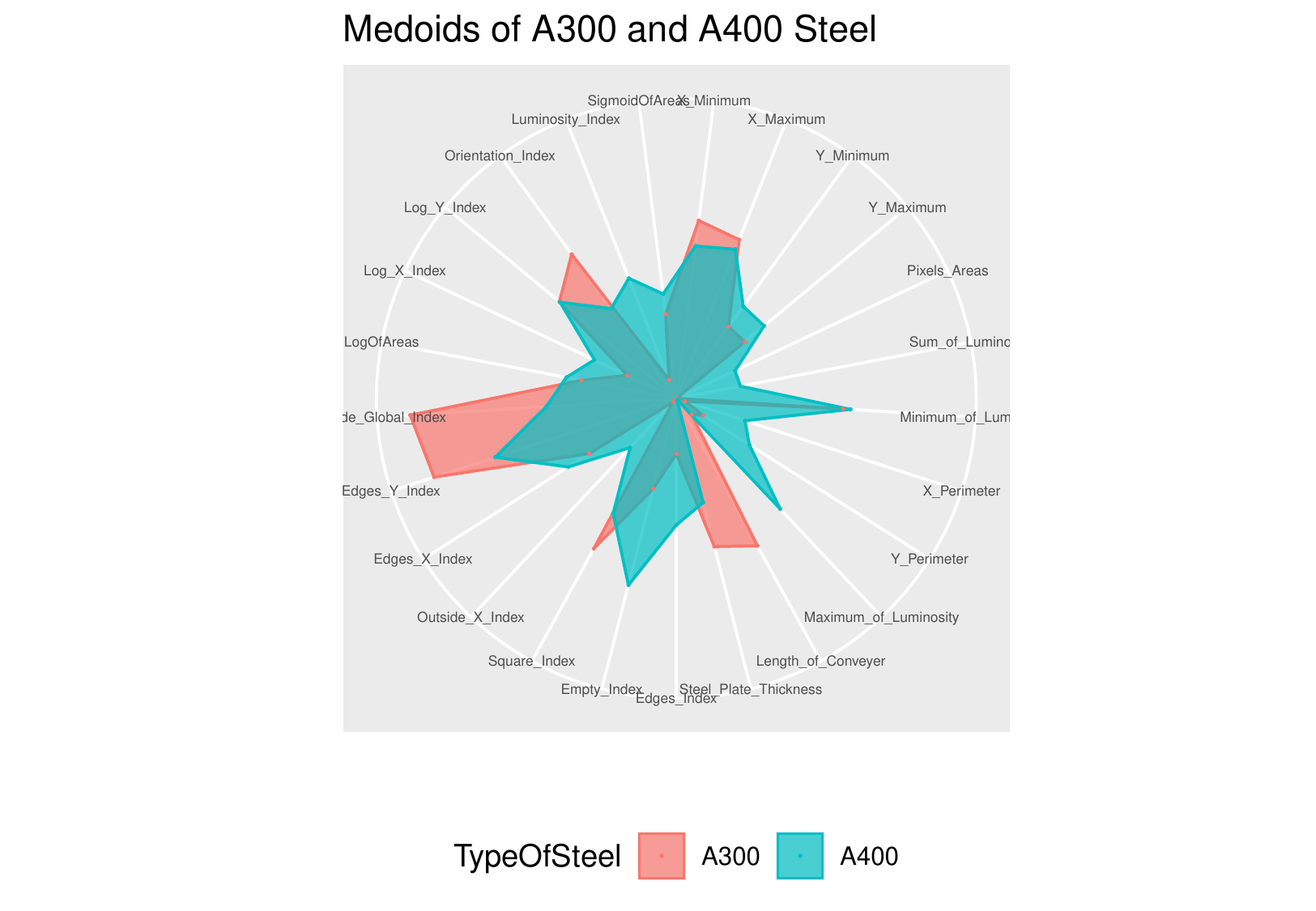} \caption{A300 and A400 have distinct footprints on the radar plot: A300 (red) is lean and covers middle to max values of quite a few variables, whereas A400 (green) is mostly concentrated at the initial (min) values of almost all variables.}\label{fig:radarA300}
\end{figure}

`Type Of Steel' has important implications for the possible Faults. Intrestingly Fault distribution for both kinds of steels i.e.~A300 and A400 is highly imbalanced e.g.~only one each instance of K\_Scratch and Stains is found in A300 type of steel. Also Dirtiness appears to be rare for this type of steel. On the other side Z\_Scratch is relatively a rare event for A400 type. Details may be further seen in Table \ref{tab:A300}. Medoids are important notion used in this work. Details shall be given latter, for the time being it is roughly defined as a prototypical observation (row of data) for a given category (say) `A300'. Medoids of A300 and A400 types of steel have been given in Figure \ref{fig:radarA300}.\\

\begin{table}[H]

\caption{\label{tab:A300}TypeOfSteel vs. Faults}
\centering
\begin{tabular}[t]{lrr}
\toprule
Fault & A300 & A400\\
\midrule
Bumps & 279 & 123\\
Common\_Other & 266 & 407\\
Z\_Scratch & 172 & 18\\
Pastry & 49 & 109\\
Dirtiness & 9 & 46\\
\addlinespace
K\_Scratch & 1 & 390\\
Stains & 1 & 71\\
Total & 777 & 1164\\
\bottomrule
\end{tabular}
\end{table}

\hypertarget{balancing}{%
\subsection{Class Balancing}\label{balancing}}

As this is an imbalanced multiclass problem class balancing has rightly been pointed by (Tian, Fu, and Wu 2015) to be a promising direction of improvement. Counts and percentages for each class are shown in the bar plot of Figure \ref{fig:imbalance}:

\begin{figure}[h]
\includegraphics{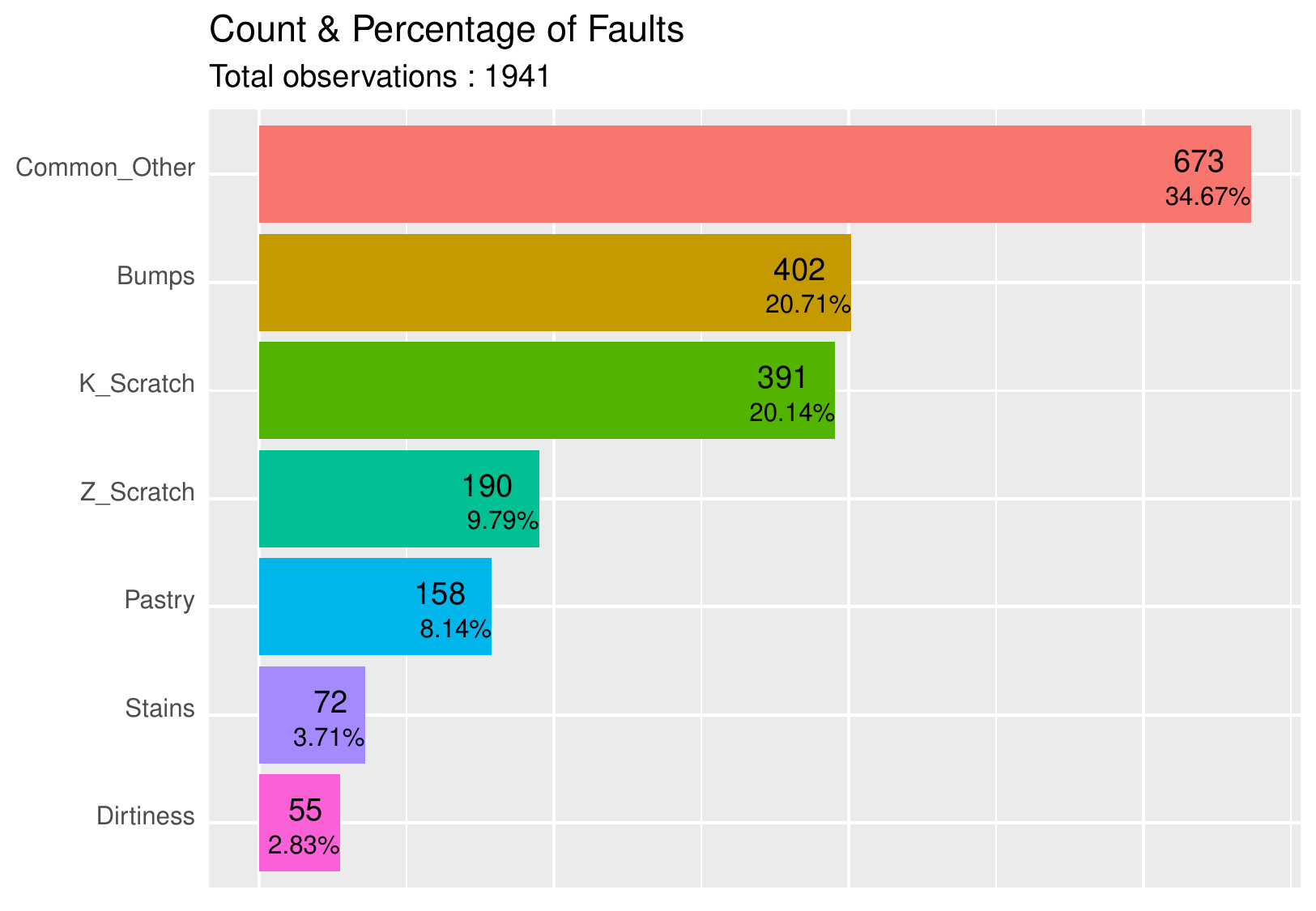} \caption{Multiclass Imbalance Classification Problem: Difference between the largest class CommonOther, and the smallest class Dirtiness, is of 618 observations. This definitely requires balancing of the dataset otherwise machine learning algorithms will be highly biased towards the majority class i.e. CommonOther }\label{fig:imbalance}
\end{figure}

As most of the machine learning algorithms carry a tacit assumption of labels of a classification dataset being equally distributed, therefore for an imbalanced problem, like the present one incorporate bias towards the majority class (Pan et al. 2020). Even if the size of dataset is large enough to provide for considerable cases of each class, the problem of validation of the generlization capability of the trained model becomes difficult. The situation even worsens for an extreme rarity condition (Kang 2020) alongwith the inhenerent randomness of data.

Various solutions and techniques have been developed to address this imbalance problem (Santiago Egea Gomez et al. 2019). Out of these three main ones have been chosen namely, undersampling, oversampling and SMOTE. Weheras the names of undersampling and oversampling render their internal working clear, technique of SMOTE (Synthetic Minority Oversampling Technique) needs some explanation. SMOTE was originally introduced in (Chawla et al. 2002) and it is basically an oversampling method but before oversampling it creats, or technically saying `synthesizes', a new instance of the minority class which is mathematically on the straight line between a pair of minority instance and one of its K nearest neighbors. This procedure is replicated for a predefined number of times.

By its very design SMOTE is implemented for a binary classification settings. However this work extended its computation for a multiclass settings using turn by turn slection of max-class and each one class from the rest of smaller classes and then applying SMOTE upon each of the binary-class splits of the original data. Using all the 3 strategies upon the original dataset in total, yielded four datasets. Characteristics of each of these datsets are given in Table \ref{tab:balanced}:

\begin{table}[H]

\caption{\label{tab:balanced}Datasets for (Im)Balancing}
\centering
\begin{tabular}[t]{lrrr}
\toprule
Dataset & Rows & Features & Class Size\\
\midrule
Original & 1941 & 27 & NA\\
Undersample & 385 & 27 & 55\\
Oversample & 4711 & 27 & 673\\
SMOTE & 4711 & 27 & 673\\
\bottomrule
\end{tabular}
\end{table}

Various machine learning algorithms have been applied to the problem of fault diagnostics of steel plates as noted in Introduction of this paper. Almost all of these methods were though aimed at achieving higher accuracy rather than gaining insight of the fault producing mechanism, as is the case here. To make an exhaustive and definitive search, present work employed 9 machine learning algorithms on each of the four datasets of Table \ref{tab:balanced}, thus resulting to 36 machine learning experiments in total. However such a large number of comparisions requires a level-field for all algorithms. Therefore a common-to-all 10 fold cross validation scheme was provided to each of the 36 ML experiments. Hardware is an HP Laptop with Intel\textsuperscript{\textregistered} Core i7-7500U CPU @ 2.70Ghz. Final results of the comparison of 9 learners over the 4 datsets are given in Table \ref{tab:benchmarking}.

\begin{table}[H]

\caption{\label{tab:benchmarking}Classification Accuracy of Learners vs. Datasets}
\centering
\resizebox{\linewidth}{!}{
\begin{tabular}[t]{lrrrr}
\toprule
Learner & Original & Oversample & SMOTE & Undersample\\
\midrule
Random Forest & 0.7831 & 0.9418 & 0.9263 & 0.7425\\
K Nearest Neighbours & 0.7285 & 0.9091 & 0.9125 & 0.7040\\
Extreme Gradient Boosting & 0.7274 & 0.8790 & 0.8588 & 0.7016\\
Support Vector Machine & 0.7537 & 0.8684 & 0.8792 & 0.7167\\
Gen. Linear Model (elastic net) & 0.7100 & 0.8400 & 0.8700 & 0.7300\\
\addlinespace
Linear Discriminant Analysis & 0.6847 & 0.7442 & 0.7739 & 0.6963\\
Naive Bayes & 0.5806 & 0.6948 & 0.7334 & 0.6651\\
Decision Tree & 0.6811 & 0.6911 & 0.7056 & 0.6651\\
Logistic Regression & 0.3936 & 0.1874 & 0.1900 & 0.1506\\
\bottomrule
\end{tabular}}
\end{table}

\hypertarget{opt}{%
\subsection{Optimization}\label{opt}}

Random Forest performed best on both the simple oversampling and SMOTE datasets. Considering that Oversampling is basically multiple copies of the minority classes and thus do not expose the randomness and internal structure of the variation of different types of faults and this work has a higher priority for extracting human-useful insights over the precision of predictions hence SMOTE data has been chosen. Hyperparameters of the Random Forest have also been optimized, with following values of the final Random Forest (henceforth RF) model; Number of trees: 186, Mtry: 5, Target node size: 1, Variable importance using: permutation, Splitrule: gini, OOB prediction error: 7.37\%. Confusion matrix for the optimized RF model is given as Table \ref{tab:confusion}, with `True' in rows and `Predicted' in Columns:

\begin{table}[H]

\caption{\label{tab:confusion}Confusion Matrix}
\centering
\resizebox{\linewidth}{!}{
\begin{tabular}[t]{lrrrrrrr}
\toprule
true & Bumps & Common\_Other & Dirtiness & K\_Scratch & Pastry & Stains & Z\_Scratch\\
\midrule
Bumps & 579 & 63 & 1 & 1 & 27 & 1 & 1\\
Common\_Other & 124 & 467 & 4 & 6 & 51 & 4 & 17\\
Dirtiness & 1 & 0 & 672 & 0 & 0 & 0 & 0\\
K\_Scratch & 3 & 12 & 0 & 656 & 1 & 1 & 0\\
Pastry & 9 & 11 & 0 & 0 & 653 & 0 & 0\\
\addlinespace
Stains & 0 & 1 & 0 & 0 & 0 & 672 & 0\\
Z\_Scratch & 1 & 3 & 0 & 0 & 0 & 0 & 669\\
\bottomrule
\end{tabular}}
\end{table}

It is notable here for later use that Bumps and Common\_Other seem to have something common in them. An `overlap' while labeling the faults may not be ruled out here. Error percentage for each type of fault as calculated from the Confusion Matrix is also shown in Table \ref{tab:confpropor}.

\begin{table}[H]

\caption{\label{tab:confpropor}Confusion Matrix : proportions}
\centering
\begin{tabular}[t]{lr}
\toprule
Fault & \%age\\
\midrule
Bumps & 15.2\\
Common\_Other & 11.8\\
Dirtiness & 14.4\\
K\_Scratch & 14.1\\
Pastry & 15.5\\
\addlinespace
Stains & 14.4\\
Z\_Scratch & 14.6\\
\bottomrule
\end{tabular}
\end{table}

\hypertarget{insights}{%
\section{Insight Harvesting}\label{insights}}

This work primarily intends to harvest insight using which it achieves higher predictive precision. Insight is the accurate and deep (in the sense of structural) understanding of a system which is rooted into external real world. This in itself remains a subject of philosophical analysis whether explanation equates to an insight. This question becomes even more pronounced in view of the consideration that explaination is about the internal working whereas insight is somewhat externalized and related to real world causal relations. By its very nature insight is obtained by infusion of various pieces of information obtained from different sources and attempts. In the same vein this work gleans various pieces of information namely XAI, Optimized Random Forest based extracted rules and Association Rule Mining (henceforth ARM). Following point made in (Christina B. Azodi, Jiliang Tang, and Shin-Han Shiu 2020) is very much in point here:\\
\emph{``Just like there is no one universally best ML algorithm, there will not likely be one ML interpretation strategy that works best on all data or for all questions. Rather, the interpretation strategy should be tailored to what one wants to learn from the ML model and confidence in the interpretation will come when multiple approaches tell the same story.''}

\hypertarget{xai}{%
\subsection{Explainable Artificial Intelligence}\label{xai}}

XAI is an umbrella term for a host of tools and techniques ranging from Mean Loss based variable importance and Model Specific tools to Model Agnostic techniques. The taxanomies introduced in (Barredo Arrieta et al. 2020) and (Baniecki and Biecek 2020) are very beneficial to the intrested reader. XAI provides for domain expert learning from ML, human requirement of automatic decisions, compliance of regulations and passing of security audits of the trained predictive models. Industrial Informatics has also flourished in recent times by using XAI (De Silva et al. 2020). Specifically present work resorts to Mean Loss based feature importance, Ceteris Peribus Profiles, Partial Dependence Profiles and Breakdown Profiles for harvesting insights into the trained model.

\hypertarget{featimp}{%
\subsubsection{Feature Importance}\label{featimp}}

\begin{figure}[H]
\includegraphics{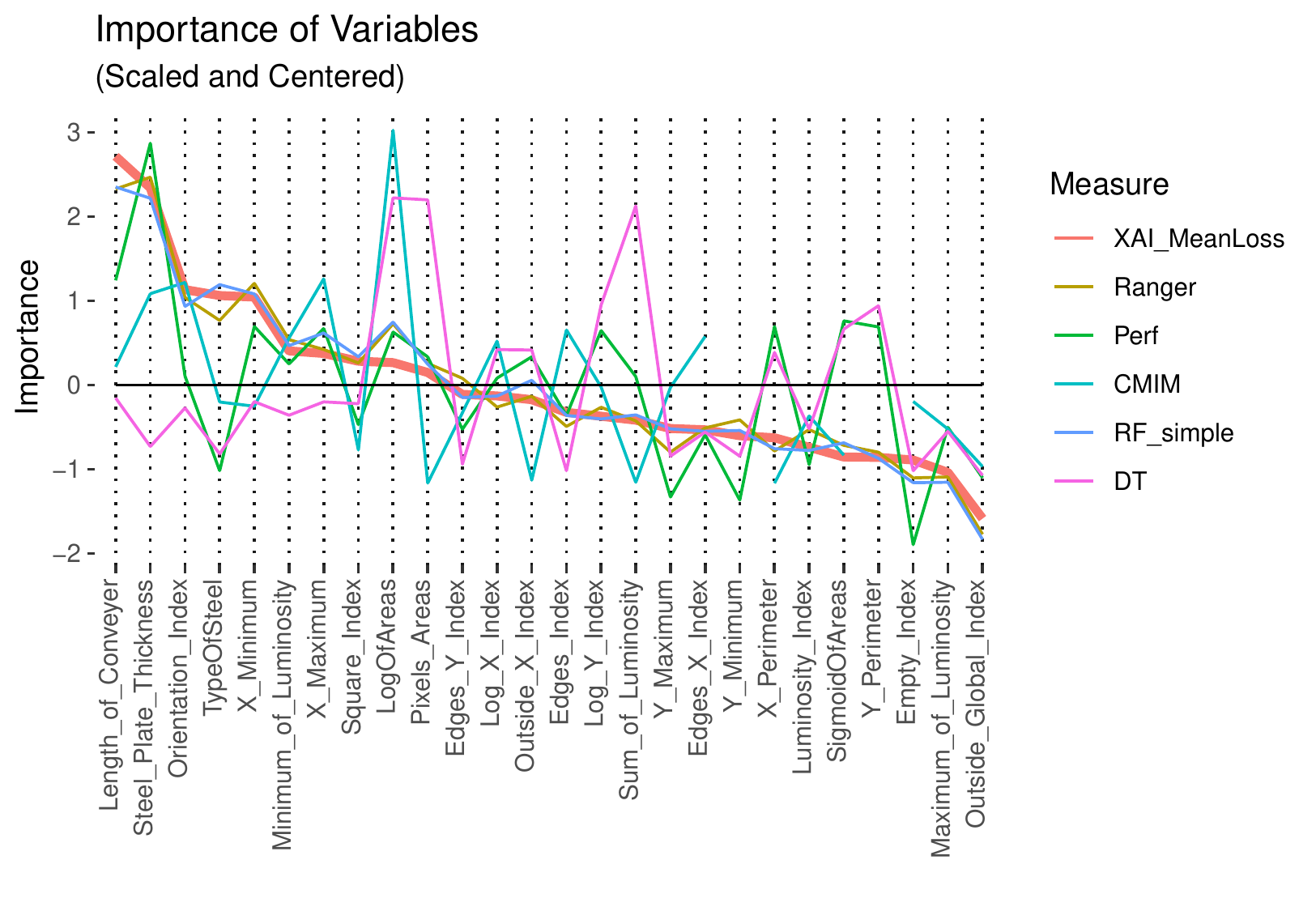} \caption{Variable Importance: Various criteria are available for importance measurement. This work mainly depends upon the Mean Loss (thick red line). All other measures have been scaled appropriately for plotting purpose.}\label{fig:varimp}
\end{figure}

Feature importance is to be kept in view throughout as a number of techniques in XAI are available for almsot all variables and thus produce unmanageable heap of information. Knowing the most significant as well as the non-colinear variables helps to reduce the cognitive load for ML interpretations. Non-colinearity is also a strong requirement in XAI tools like PDP. Hence computation of feature importance is the natural first step.\\
Besides the feature importances obtained from the optimized Random Forest, a number of other measures may be computed and combined to benchmark the features for subsequent use (Figure \ref{fig:varimp}). Short description of each computed measure is given as:\\
- CMIM: Minimal Conditional Mutual Information\\
- DT: Decision Tree\\
- Performance: Predictive Performance\\
- Ranger: RF Optimized using permutation strategy\\
- RF\_simple: Random Forest simple\\
- Mean Loss: model agnostic XAI technique

Table \ref{tab:varimptbl} shows the importance ranks of each feature (with 1 having largest mean loss). This importance rank shall be used subsequently to figure out their relative places within the Breakdown plots of each fault-medoid.\\

\begin{table}[H]

\caption{\label{tab:varimptbl}Importance Ranks of Variables}
\centering
\begin{tabular}[t]{llll}
\toprule
Feature & Rank & Feature & Rank\\
\midrule
Length\_of\_Conveyer & 1 & Edges\_Index & 14\\
Steel\_Plate\_Thickness & 2 & Log\_Y\_Index & 15\\
Orientation\_Index & 3 & Sum\_of\_Luminosity & 16\\
TypeOfSteel & 4 & Y\_Maximum & 17\\
X\_Minimum & 5 & Edges\_X\_Index & 18\\
\addlinespace
Minimum\_of\_Luminosity & 6 & Y\_Minimum & 19\\
X\_Maximum & 7 & X\_Perimeter & 20\\
Square\_Index & 8 & Luminosity\_Index & 21\\
LogOfAreas & 9 & SigmoidOfAreas & 22\\
Pixels\_Areas & 10 & Y\_Perimeter & 23\\
\addlinespace
Edges\_Y\_Index & 11 & Empty\_Index & 24\\
Log\_X\_Index & 12 & Maximum\_of\_Luminosity & 25\\
Outside\_X\_Index & 13 & Outside\_Global\_Index & 26\\
\bottomrule
\end{tabular}
\end{table}

Most of the XAI tools work best only for the least correlated features with no or minimum colinearity. Therefore correlation coefficient of all features has been calculated for filtering of features and five features namely, Sum\_of\_Luminosity, Pixels\_Areas, X\_Perimeter, X\_Minimum, Y\_Minimum are having correlation more than 0.90. It is noted here for subsequent usage in XAI tools where needed e.g.~in Ceteris Peribus profiles and Breakdown plots.

\hypertarget{cp}{%
\subsubsection{Ceteris Paribus for Fault Medoids}\label{cp}}

A model which is otherwise opaque, may easily be peeped into if effect of each explanatory variable is examined separately. A further improvement is to examine one variable while keeping all others at some constant value, so to say, average value. Ceteris Peribus (CP) profile is the tool for this kind of analysis. Ceteris Peribus is latin meaning ``all others being constant''. CP investigates the local curvature of the response surface of the black-box model (Goldstein et al. 2015). From the insight viewpoint it provides an indication how, in actual world, a variable may be affecting the fault creation. In literature CP technique has also been named as `what-if' analysis or Individual Conditional Expectations (ICE).

CP may be rigorously defined as follows: Let \(D\) stand for a dataset with \(n\) rows and \(p\) columns. Here \(p\) stands for the number of variables while \(n\) stands for the number of observations. As local methods operate on a single observation, let \(x^{\ast }\in \mathbb{R}\) stand for an observation of interest. Let \(f:X\rightarrow \mathbb{R}\) denote for the model of interest, where \(X=\mathbb{R}^{p}\) is the \(p\)-dimensional input space. Then Ceteris Paribus is the profile \(g\left(z\right)\) for variable \(x_{i}\) and observation \(x\) is defined as:
\[
g_{x^{\ast }}(z)=f(x^{\ast }|x_{i}=z)
\]

As CP is a local technique i.e.~based upon a single instance (row) of data. Thus any row in data may be chosen, however it is more suitable for the current scenario to choose to represent each Fault type by its most representative instance. In cluster analysis an instance with minimum dissimilarity (Yu et al. 2018) within a cluster is said to be its medoid. This notion is akin to taht of `centroid' in computation of clusters. While various methods of medoid calculation are available we choose the fastest one i.e.~medians of all numeric variables and mode (most frequent level) of a categorical variable. It is important to note that medoid here have been calculated from the original imbalanced dataset so that they remain `typical' and no randomness creeps in. So to say, for this work a medoid refers to an object within a Fault-set for which average dissimilarity between it and all the other members of the cluster is minimal. Other such works demonstrating the usefulness of typical prototypes for XAI may be found in (Gurumoorthy et al. 2019) and the refernces therein.\\
Typical fault profiles (medoids) used herein are given as Table \ref{tab:medoids} ordered by Variable Importance ranks as given in Table \ref{tab:varimptbl}:

\begin{table}[H]

\caption{\label{tab:medoids}Medoids (prototypical examplars) for Different Faults. Ordered by Mean Loss Based Feature Importance}
\centering
\resizebox{\linewidth}{!}{
\begin{tabular}[t]{llllllll}
\toprule
Variable & Pastry & Z\_Scratch & K\_Scratch & Stains & Dirtiness & Bumps & Common\_Other\\
\midrule
Length\_of\_Conveyer & 1648 & 1356 & 1362 & 1358 & 1364 & 1624 & 1372\\
Steel\_Plate\_Thickness & 85 & 70 & 40 & 50 & 100 & 70 & 70\\
Orientation\_Index & 0.66670 & 0.35145 & -0.51520 & -0.26785 & 0.80000 & 0.10000 & 0.18180\\
TypeOfSteel & A300 & A300 & A400 & A400 & A400 & A300 & A400\\
X\_Minimum & 823.0 & 106.0 & 41.0 & 754.0 & 625.0 & 856.5 & 655.0\\
\addlinespace
Minimum\_of\_Luminosity & 80 & 93 & 41 & 115 & 107 & 91 & 97\\
X\_Maximum & 836.0 & 123.0 & 212.0 & 760.5 & 636.0 & 878.0 & 727.0\\
Square\_Index & 0.3333 & 0.5833 & 0.4637 & 0.7033 & 0.1936 & 0.7500 & 0.5333\\
LogOfAreas & 2.32010 & 2.16435 & 3.79800 & 1.21730 & 2.16140 & 2.08100 & 2.16440\\
Pixels\_Areas & 209.0 & 146.0 & 6281.0 & 16.5 & 145.0 & 120.5 & 146.0\\
\addlinespace
Edges\_Y\_Index & 1.0000 & 0.9310 & 0.4744 & 1.0000 & 1.0000 & 1.0000 & 0.9375\\
Log\_X\_Index & 1.00000 & 1.17610 & 2.19030 & 0.77820 & 1.00000 & 1.09655 & 1.14610\\
Outside\_X\_Index & 0.00660 & 0.01105 & 0.11270 & 0.00440 & 0.00730 & 0.00810 & 0.01010\\
Edges\_Index & 0.15015 & 0.15235 & 0.05860 & 0.59015 & 0.89740 & 0.44150 & 0.32060\\
Log\_Y\_Index & 1.48425 & 1.30100 & 1.82610 & 0.60210 & 1.54410 & 1.14610 & 1.25530\\
\addlinespace
Sum\_of\_Luminosity & 21342.0 & 16102.0 & 654358.0 & 2124.0 & 17271.0 & 12819.5 & 16987.0\\
Y\_Maximum & 1079078 & 1236224 & 1512467 & 1133486 & 2046183 & 1543061 & 880535\\
Edges\_X\_Index & 0.5000 & 0.5000 & 0.6189 & 1.0000 & 0.2903 & 0.7222 & 0.6471\\
Y\_Minimum & 1079061 & 1236179 & 1512277 & 1133482 & 2046094 & 1543048 & 880521\\
X\_Perimeter & 20 & 28 & 271 & 7 & 28 & 17 & 25\\
\addlinespace
Luminosity\_Index & -0.17905 & -0.16455 & -0.15630 & -0.00255 & -0.08840 & -0.14305 & -0.11880\\
SigmoidOfAreas & 0.51125 & 0.45720 & 1.00000 & 0.13840 & 0.40570 & 0.30970 & 0.41700\\
Y\_Perimeter & 30.5 & 23.0 & 139.0 & 4.5 & 35.0 & 16.0 & 22.0\\
Empty\_Index & 0.32070 & 0.45935 & 0.45680 & 0.40835 & 0.39900 & 0.33330 & 0.42310\\
Maximum\_of\_Luminosity & 126 & 124 & 134 & 140 & 132 & 127 & 127\\
\addlinespace
Outside\_Global\_Index & 1 & 1 & 0 & 0 & 1 & 1 & 1\\
\bottomrule
\end{tabular}}
\end{table}

\begin{figure}[h]
\includegraphics{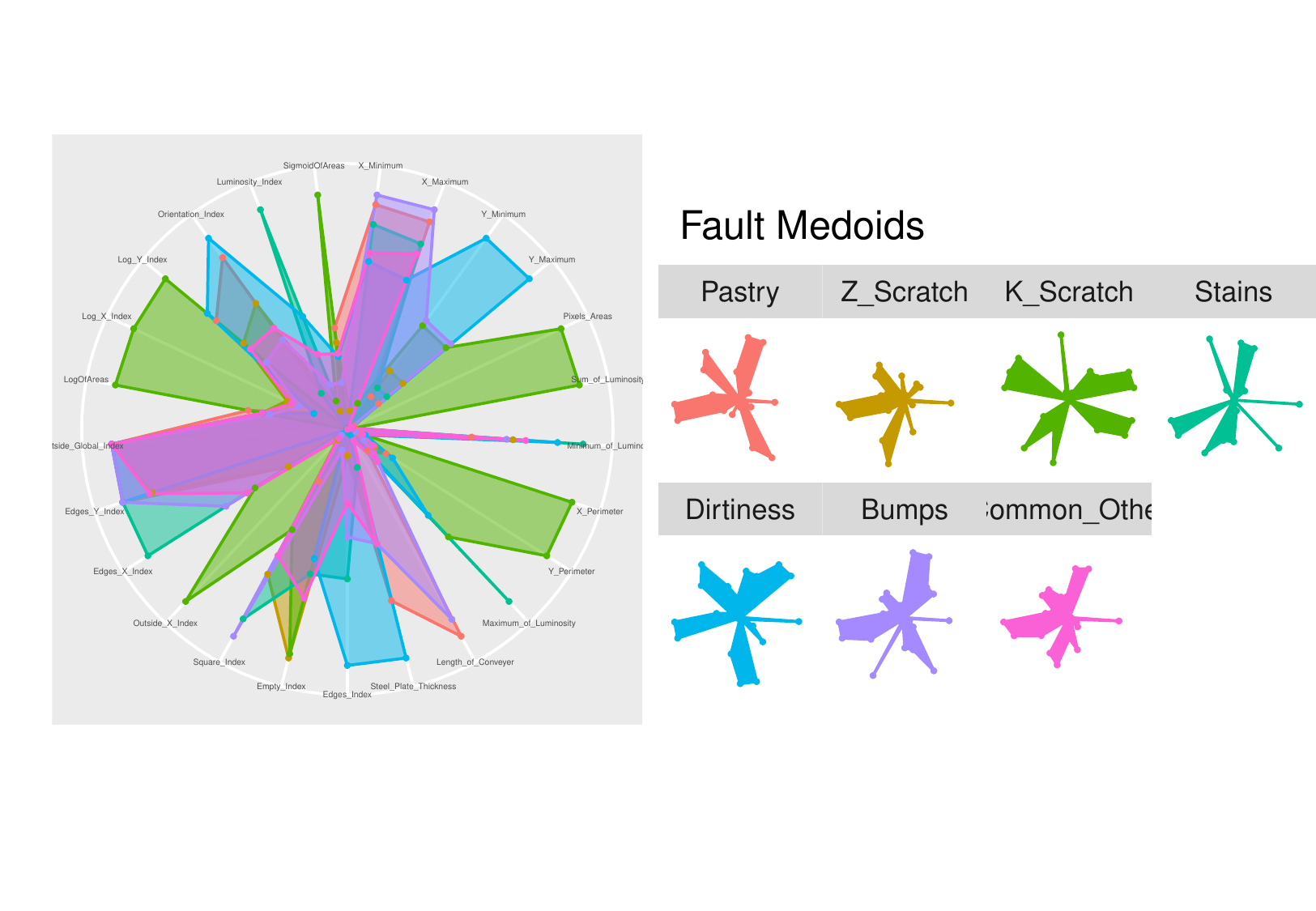} \caption{Radar Plot(s) of Medoids: Smaller radars present the profile of each medoid individually and at the same time they are legend for the combined radar plot at left.}\label{fig:medoidradar}
\end{figure}

Once each numeric variable is scaled, these medoids may also be visualized as radar plots (Figure \ref{fig:medoidradar}). A closer examination of the radar patterns reveals following insights:

\emph{Insights}\\
1. Bumps and Common\_Other are having striking similarity of shape with each other.\\
2. Z\_Scratch \& K\_Scratch are very distinct and their mutual defferetiation is easier.\\
3. A typical Dirtiness case (i.e.~medoid) is spread almost all over the 26 features.\\
4. Except K\_Scratch all other typical fault cases have presence on the `Minimum\_Of\_Luminosity' feature. Visually it may be see as a tail in the radar plot. This renders this variable important for classification.

\begin{figure}[h]
\includegraphics{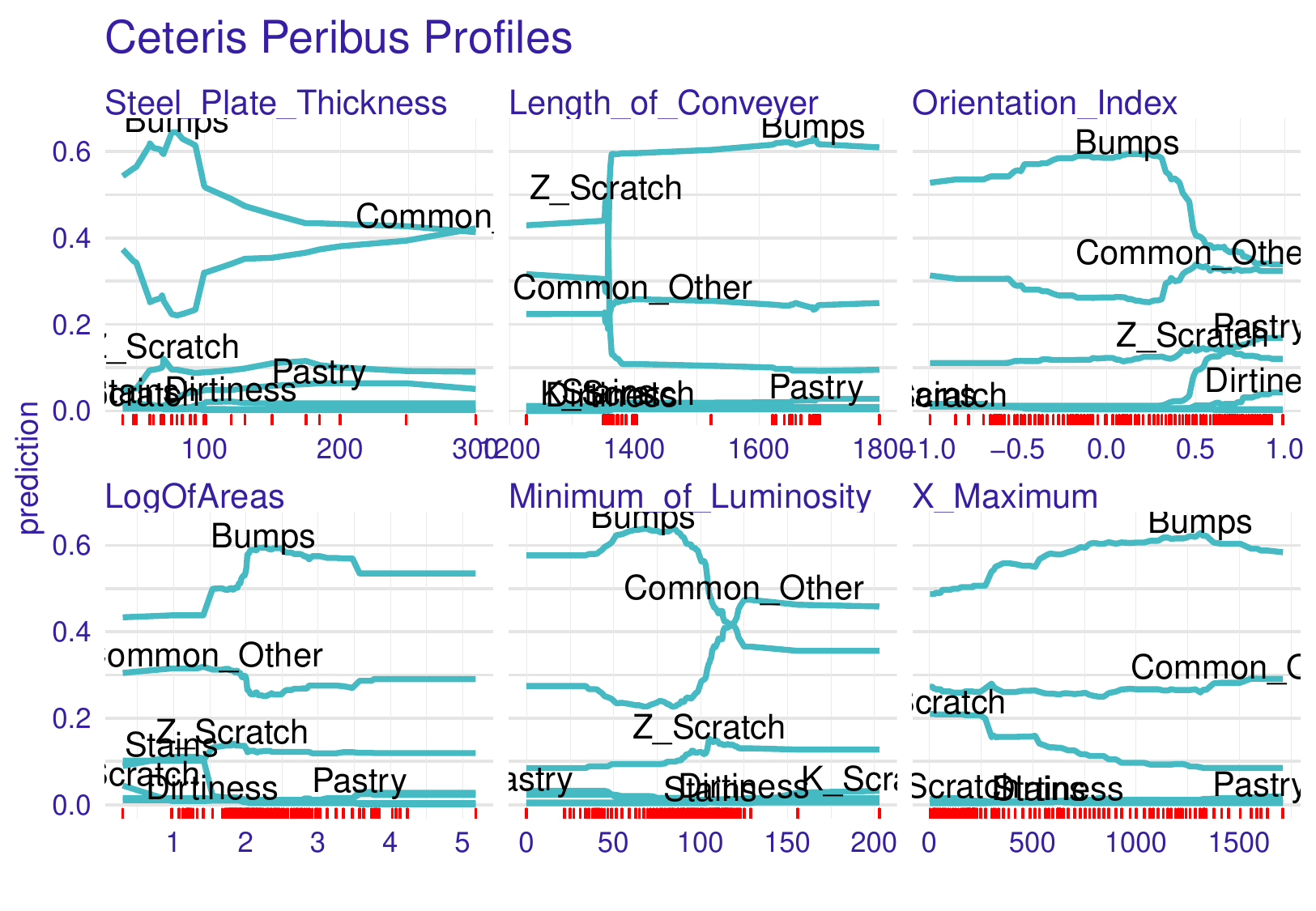} \caption{Ceteris Peribus Profiles: Although CP is essentially a local (i.e. limited to the single observation of intrest) technique, however here the localness syndrom has been balanced by the use of Fault Medoids. Density rugs on the horizontal axis provide an indication of strength of available observations equivalently strength of evidence for the CP patterns.}\label{fig:cpProfiles}
\end{figure}

Figure \ref{fig:cpProfiles} provides Ceteris Peribus (CP) profiles of only 6 important variables because for each of the seven medoids and 6 variables these turns out to be 42 CP-plots. The real utility of CP profile shines once the insights, general observing of the similarities, regularities, pecularities and/or differences, are noted and separated for later use by human experts or by the machine algorithms. Therefore for the variables not shown here, just the gleaned insights are listed below:

\emph{Insights}\\
1. Bumps and Common\_Other almost always mirror each other in CP profile of all variables except X\_Maximum. This indicates a kind of `overlap' while labeling has been originally done.\\
2. For thinner plates (with thickness less than 100) maximum probablity is for Bumps, which is understandabale.\\
3. As the density rugs provide indication of support (by number of available observations), therefore e.g.~Steel\_Plate\_Thickness is `intresting' for values less than 100. Beyond that point curves are flat and hence unintresring. Moreover beyond 100 supporting observations are scarce thus weakening the confidence of prediction.

\hypertarget{pdp}{%
\subsubsection{Partial Dependence Plots}\label{pdp}}

Partial Dependence (PD) Plot describes how certain set of variables affect an average prediction. It is important to differentiate between Ceteris Peribus (CP) and PD. CP is localized to a certain instance, and in present work this happens to be medoid of a fault. PD is non-local in its character as it talks about a general `average' prediction, albiet the set of variables being chosen by the user. PD shows how the prediction `partially' depends upon the few variables of intrest. PD also describes the kind of relationaship e.g.~linear, cuvilinear and step etc. More elaborately, a PDP examines a variable of intrest at a specific range. At each value within the range, the predictive model predicts for all cases (rows), and the prediction is then averaged out. Therefore a relation depicted by PDP is useful only once the features are non-colinear.\\
Technically, a PD may be written (Friedman 2001) as:
\[
\hat{f}_{x_{s}}\left( x_{S}\right) =E_{x_{C}}\left[ \hat{f}%
_{x_{s}}\left( x_{S},x_{C}\right) \right]=\int \hat{f}_{x_{s}}\left(
x_{S},x_{C}\right) d\mathbb{P}\left( x_{C}\right)
\]

where \(x_{S}\) are the features for which the partial dependence function
should be plotted and \(x_{C}\) are the other features used in the machine
learning model \(\hat{f}\).

The partial function \(\hat{f}_{x_{s}}\) is estimated by calculating
averages in the training data, also known as Monte Carlo method:
\[
\hat{f}_{x_{s}}\left( x_{S}\right) =\frac{1}{n}\sum\limits_{i=1}^{n}%
\hat{f}_{x_{s}}\left( x_{S},x_{C}^{\left( i\right) }\right)
\]

\begin{figure}[h]
\includegraphics{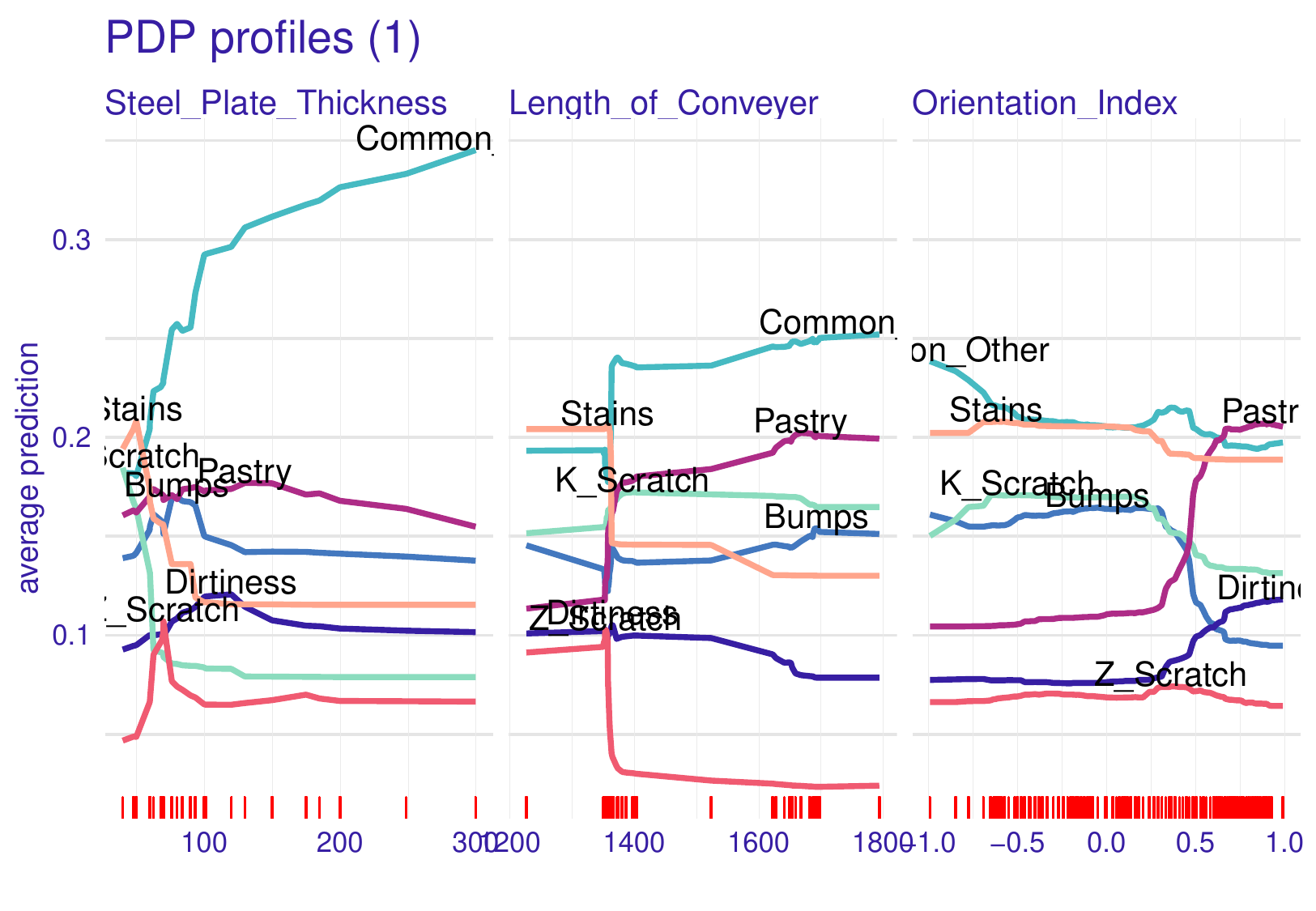} \caption{Partial Dependence Plots - 1: Dependence of an average prediction upon different variables. Contribution of Steel Plate Thickness to CommonOther type of fault is noteworthy.}\label{fig:pdpProfiles1}
\end{figure}

\begin{figure}[h]
\includegraphics{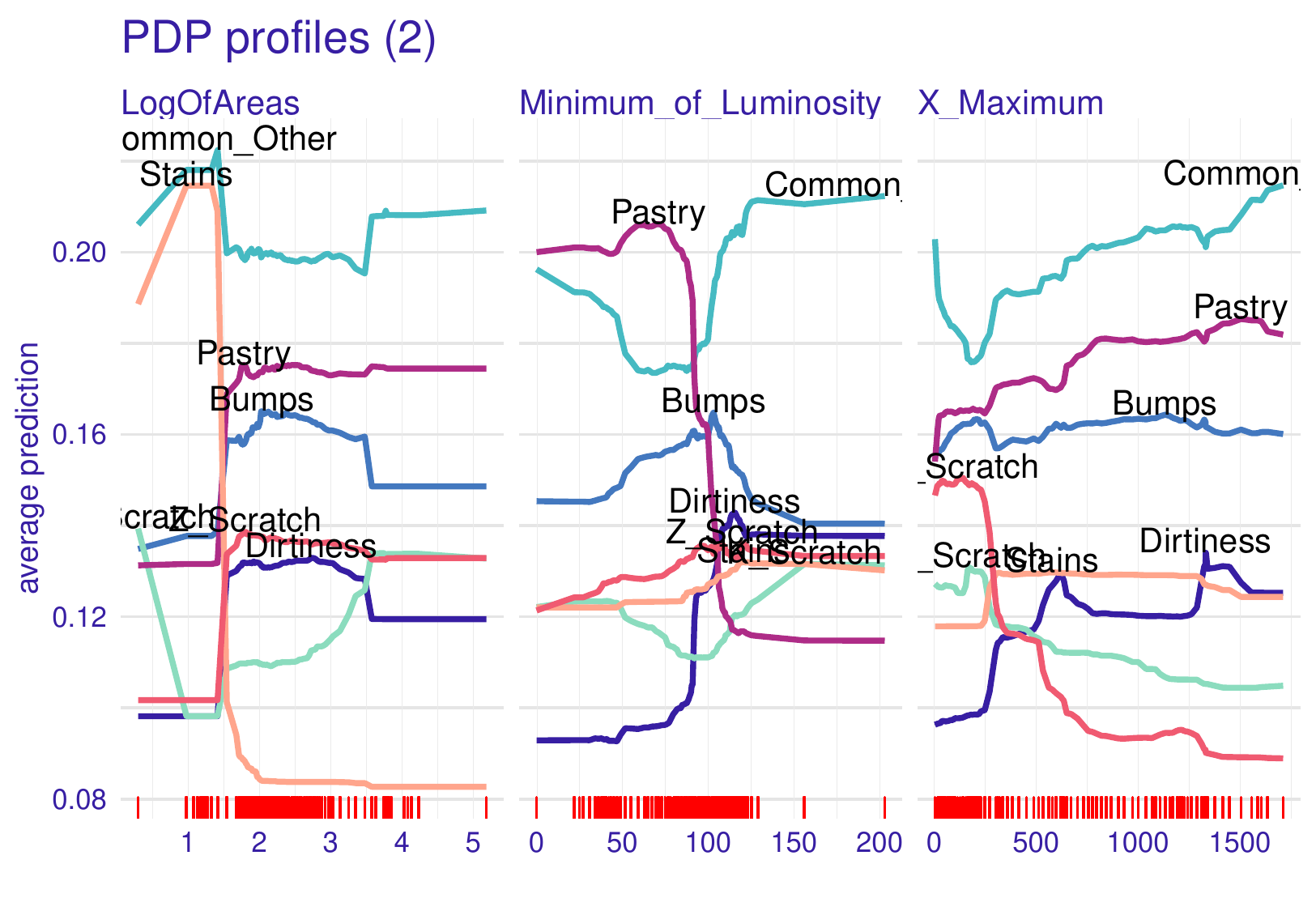} \caption{Partial Dependence Plots - 2: Horizontal density rugs in red color hold the same significance as that in CP plot earlier.}\label{fig:pdpProfiles2}
\end{figure}

\begin{figure}[h]
\includegraphics{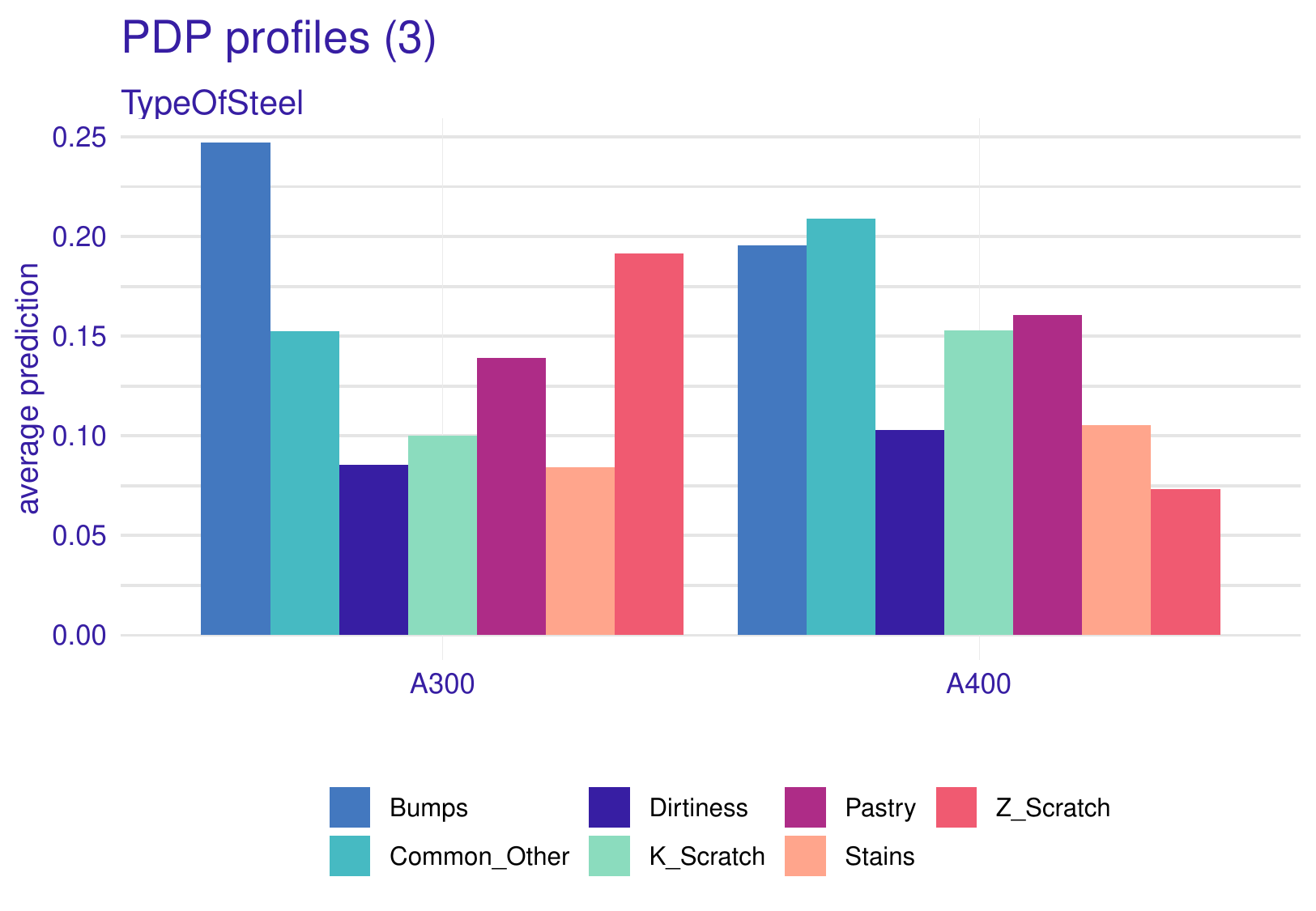} \caption{Partial Dependence Plots - 3: A300 is more prone to Bumps type of fault. PDP is given in the form of histogram as the variable of intrest is a categorical one instead of the prvious two PDP charts were variables are of numeric type.}\label{fig:pdpProfiles3}
\end{figure}

This work, in the manner of (R. Berk and J. Bleich 2013), demonstrates the benefit of using optimized Random Forest and its PDPs to precisely indicate the predictor-response relationships under imbalanced classification settings which are quite common in industrial problems.\\
As humans can percieve only a few number of variables at any given time therefore in Figure \ref{fig:pdpProfiles1}, \ref{fig:pdpProfiles2} and \ref{fig:pdpProfiles3} target features for PD plots are chosen to be small in number and those also from the most important variables/features.

\hypertarget{bd}{%
\subsubsection{Breakdown Profiles}\label{bd}}

\begin{figure}[h]
\includegraphics{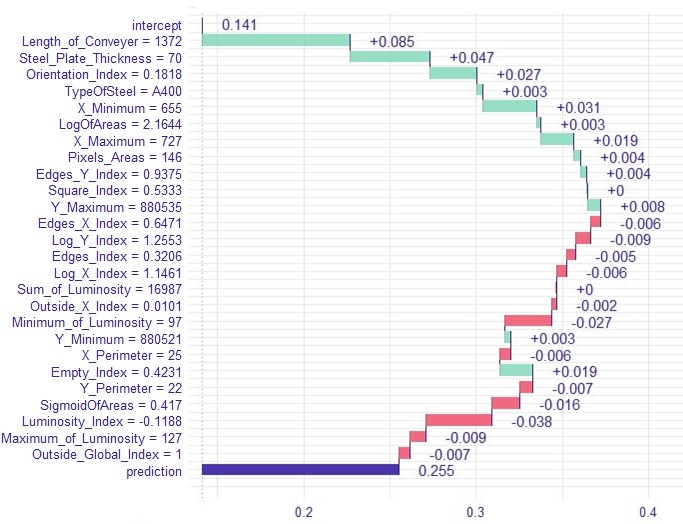} \caption{Breakdown Plot for CommonOther medoid. Breakdown portrays the contribution (positive/negative) of each predictive feature towards the final prediction of a certain observations, which, in this case is a medoid.}\label{fig:BDplot}
\end{figure}

Medoids of each fault provides an important opportunity to figure out a hypothesis of how different variables affect the model-prediction of the specific medoid. Thus indicating the significance of different variables for different types of predictions. This is a kind of variable attribution exercise which attributes contribution of each individual variable to the present result. Technically known as Breakdown (BD) Profiles, there full plots showing each of the 27 variables is too space consuimg. Therefore only one BD plot is shown in Figure \ref{fig:BDplot}. One easily finds the most important contributors from a visual analysis of this plot. The variables in red color drag the prediction `backward' and positive contributors push it forward and hence the net prediction result (in deep blue) is obtained.

For rest of the Fault-medoids only top-10 bigger contributors are listed alongwith their positive or negative contributions (using the +/- sign). For saving space we present the top ten features from the Breakdown plot of each fault-medoid in the following:

\emph{Bumps}\\
Empty\_Index +0.095, Square\_Index +0.068, Luminosity\_Index +0.056, TypeOfSteel +0.047, Orientation\_Index +0.046, Length\_of\_Conveyer +0.045, Edges\_X\_Index +0.042, X\_Perimeter +0.038, Minimum\_of\_Luminosity +0.038, Maximum\_of\_Luminosity +0.03

\emph{Common\_Other}\\
Length\_of\_Conveyer +0.085, Steel\_Plate\_Thickness +0.052, X\_Minimum +0.033, Luminosity\_Index -0.027, Minimum\_of\_Luminosity -0.027, Orientation\_Index +0.027, Edges\_X\_Index -0.025, Edges\_Y\_Index +0.021, X\_Maximum +0.018, TypeOfSteel +0.012

\emph{Dirtiness}\\
Y\_Maximum +0.132, Y\_Minimum +0.121, Edges\_X\_Index +0.043, Edges\_Index +0.037, X\_Maximum +0.027, Maximum\_of\_Luminosity +0.024, TypeOfSteel -0.023, Minimum\_of\_Luminosity +0.02, Luminosity\_Index +0.016, Outside\_X\_Index +0.016

\emph{K\_Scratch}\\
Log\_Y\_Index +0.109, X\_Minimum +0.07, Sum\_of\_Luminosity +0.058, X\_Perimeter +0.05, TypeOfSteel -0.05, Pixels\_Areas +0.05, Steel\_Plate\_Thickness +0.045, X\_Maximum +0.034, LogOfAreas +0.033, SigmoidOfAreas +0.032

\emph{Pastry}\\
Edges\_Y\_Index +0.091, Empty\_Index +0.059, Orientation\_Index +0.054, Outside\_X\_Index +0.051, Y\_Perimeter +0.046, Length\_of\_Conveyer +0.044, X\_Perimeter +0.044, Luminosity\_Index +0.039, Minimum\_of\_Luminosity +0.036, Edges\_X\_Index +0.033

\emph{Stains}\\
Length\_of\_Conveyer +0.1, Steel\_Plate\_Thickness +0.089, LogOfAreas +0.075, Pixels\_Areas +0.074, Sum\_of\_Luminosity +0.065, SigmoidOfAreas +0.049, TypeOfSteel -0.048, Y\_Perimeter +0.045, Log\_Y\_Index +0.029, X\_Maximum +0.026

\emph{Z\_Scratch}\\
Length\_of\_Conveyer -0.101, TypeOfSteel +0.014, Empty\_Index -0.01, X\_Maximum -0.008, X\_Minimum -0.007, Outside\_X\_Index -0.007, Steel\_Plate\_Thickness +0.006, Luminosity\_Index -0.006, Log\_Y\_Index -0.005, Log\_X\_Index -0.004

\hypertarget{rf}{%
\subsection{Random Forest Rules}\label{rf}}

In search of greater explanability and interpretability a natural target is to search under the hood of a relatively high precision predictive model as is the case with presnt dataset and Random Forest (RF) model developed and optimized earlier. Use of variable selection for a minimal and relevent set of conditions from the association rules has been enunciated by (Deng et al. 2014). Therefore exposing of the internal structure of RF model has been carried out in line with the works (Deng et al. 2014) and (Deng 2019).

Rule based models are very good at interpretability across a wide area of applications. Such models base upon a simplified uderstanding of their internal workings. However one of the major problems with rule generation is the large number of rules besides their coverage and specificity. In simpler terms coverage is the amount and specificity is the length of a rule. Usage of these measures of rule quality is fundamentally dictated by the intention of the user. For the sake of explainability a smaller number of rules with as much coverage as possible is a pre-requisite. A rule having very lengthy antecedent or consequent becomes very difficult to understand and interpret. Therefore transparency and explainability of a model is directly proportional to the coverage and specificity of the rule-base.

In literature a large number of research works may be found regarding production of rules but considerably fewer works are seen for filtering and pruning strategies of the rulebase. This work proposes a novel approcah for model building using rules mining: discovering a large number of rules and then focusing-in the rulebase using apriori expertise e.g.~semantic analysis and the quantitative filtering on the basis of various quality measures of rules.

\begin{table}[H]

\caption{\label{tab:unnamed-chunk-4}Extracted Rules from Optimized Random Forest}
\centering
\resizebox{\linewidth}{!}{
\begin{tabular}[t]{l>{\raggedright\arraybackslash}p{5in}lrrr}
\toprule
ID & condition & pred & impRRF & err & freq\\
\midrule
RF-1 & Y\_Perimeter<=14 \& Steel\_Plate\_Thickness<=50.218 \& Steel\_Plate\_Thickness>49.827 & Stains & 1.0000000 & 0.01 & 0.14\\
RF-2 & Sum\_of\_Luminosity>146830.76 \& Orientation\_Index<=0.484 \& TypeOfSteel \%in\% c('A400') & K\_Scratch & 0.8337620 & 0.02 & 0.12\\
RF-3 & X\_Maximum>278.157 \& Y\_Minimum>1541367 \& Y\_Maximum<=2427567.836 \& Minimum\_of\_Luminosity>91.926 \& Length\_of\_Conveyer<=1489.939 \& Outside\_Global\_Index>0.977 & Dirtiness & 0.6455204 & 0.04 & 0.11\\
RF-4 & Length\_of\_Conveyer<=1358.181 \& Steel\_Plate\_Thickness<=74.543 \& TypeOfSteel \%in\% c('A300') & Z\_Scratch & 0.7992496 & 0.07 & 0.14\\
RF-5 & X\_Maximum<=98.987 \& Edges\_X\_Index<=0.543 \& Luminosity\_Index>-0.066 \& TypeOfSteel \%in\% c('A300') & Pastry & 0.0516566 & 0.09 & 0.01\\
\addlinespace
RF-6 & X\_Minimum>250.5 \& Sum\_of\_Luminosity>3463.95 \& Minimum\_of\_Luminosity>100.5 & Dirtiness & 0.0107497 & 0.56 & 0.20\\
\bottomrule
\end{tabular}}
\end{table}

The rules produced by machine are of course mechanistic and may be `humanized' by re-writing them in near human-language.

\hypertarget{arm}{%
\subsection{Association Rules}\label{arm}}

Association rule mining is now a well established sub-area of machine learning with the additional advantage of being fully transparent from the explainability view point (Grabot 2020).\\

\begin{table}[H]

\caption{\label{tab:rulesListing}Association Rules for Faults}
\centering
\resizebox{\linewidth}{!}{
\begin{tabular}[t]{llrrrr}
\toprule
ID & Rules & Confidence & Count & Lift & Support\\
\midrule
Bumps-1 & Y\_Minimum=[2.94e+06,5.56e+06),Steel\_Plate\_Thickness=[79.86,80.07) & 0.9464286 & 53 & 6.625000 & 0.0112503\\
Bumps-2 & Y\_Maximum=[2.94e+06,5.56e+06),Steel\_Plate\_Thickness=[79.86,80.07) & 0.9464286 & 53 & 6.625000 & 0.0112503\\
Bumps-3 & Length\_of\_Conveyer=[1687.97,1694.06),Log\_X\_Index=[1.079,1.202),TypeOfSteel=A300 & 0.9122807 & 52 & 6.385965 & 0.0110380\\
Bumps-4 & Empty\_Index=[0.252,0.329),Log\_X\_Index=[1.079,1.202),SigmoidOfAreas=[0.202,0.445),TypeOfSteel=A300 & 0.8928571 & 50 & 6.250000 & 0.0106135\\
Bumps-5 & X\_Perimeter=[14.03,17),Sum\_of\_Luminosity=[8.8e+03,1.3e+04),Log\_X\_Index=[1.079,1.202) & 0.8888889 & 48 & 6.222222 & 0.0101889\\
\addlinespace
Bumps-6 & Steel\_Plate\_Thickness=[79.86,80.07),SigmoidOfAreas=[0.445,0.978) & 0.8857143 & 62 & 6.200000 & 0.0131607\\
Bumps-7 & Length\_of\_Conveyer=[1687.97,1694.06),Square\_Index=[0.85,0.924),TypeOfSteel=A300 & 0.8852459 & 54 & 6.196721 & 0.0114625\\
Bumps-8 & X\_Perimeter=[14.03,17),Sum\_of\_Luminosity=[8.8e+03,1.3e+04),Edges\_Y\_Index=[1, Inf],Log\_Y\_Index=[0.952,1.15) & 0.8813559 & 52 & 6.169492 & 0.0110380\\
Bumps-9 & Length\_of\_Conveyer=[1687.97,1694.06),Steel\_Plate\_Thickness=[59.99,60.05) & 0.8709677 & 81 & 6.096774 & 0.0171938\\
Bumps-10 & Length\_of\_Conveyer=[1687.97,1694.06),Log\_X\_Index=[1.079,1.202),SigmoidOfAreas=[0.202,0.445) & 0.8596491 & 49 & 6.017544 & 0.0104012\\
\addlinespace
Dirty-1 & Y\_Maximum=[1.96e+06,2.24e+06),Length\_of\_Conveyer=[1364,1366.01) & 0.9903846 & 309 & 6.932692 & 0.0655912\\
Dirty-2 & Y\_Minimum=[1.96e+06,2.24e+06),Length\_of\_Conveyer=[1364,1366.01) & 0.9903846 & 309 & 6.932692 & 0.0655912\\
Dirty-3 & Y\_Maximum=[1.96e+06,2.24e+06),Steel\_Plate\_Thickness=[99.97,120.6) & 0.9649123 & 330 & 6.754386 & 0.0700488\\
Dirty-4 & Y\_Minimum=[1.96e+06,2.24e+06),Steel\_Plate\_Thickness=[99.97,120.6) & 0.9649123 & 330 & 6.754386 & 0.0700488\\
Dirty-5 & Length\_of\_Conveyer=[1364,1366.01),Steel\_Plate\_Thickness=[99.97,120.6) & 0.9630769 & 313 & 6.741538 & 0.0664402\\
\addlinespace
Dirty-6 & X\_Minimum=[612,625) & 0.9398281 & 328 & 6.578797 & 0.0696243\\
Dirty-7 & X\_Maximum=[622,640) & 0.9217877 & 330 & 6.452514 & 0.0700488\\
Dirty-8 & Steel\_Plate\_Thickness=[99.97,120.6),Outside\_Global\_Index=[0.997, Inf],TypeOfSteel=A400 & 0.9149425 & 398 & 6.404598 & 0.0844831\\
Dirty-9 & Length\_of\_Conveyer=[1364,1366.01),Outside\_Global\_Index=[0.997, Inf],TypeOfSteel=A400 & 0.9020173 & 313 & 6.314121 & 0.0664402\\
Dirty-10 & Edges\_Index=[0.897,0.917) & 0.9013699 & 329 & 6.309589 & 0.0698366\\
\addlinespace
KScratch-1 & Outside\_X\_Index=[0.0866,0.142) & 0.9726562 & 498 & 6.808594 & 0.1057100\\
KScratch-2 & Log\_X\_Index=[2.063,2.353) & 0.9718574 & 518 & 6.803002 & 0.1099554\\
KScratch-3 & Steel\_Plate\_Thickness=[-Inf,40.14),SigmoidOfAreas=[1, Inf] & 0.9501661 & 572 & 6.651163 & 0.1214180\\
KScratch-4 & LogOfAreas=[3.73, Inf] & 0.9480249 & 456 & 6.636175 & 0.0967947\\
KScratch-5 & Pixels\_Areas=[5.56e+03, Inf] & 0.9471459 & 448 & 6.630021 & 0.0950966\\
\addlinespace
KScratch-6 & X\_Maximum=[172,227),SigmoidOfAreas=[1, Inf] & 0.9449541 & 515 & 6.614679 & 0.1093186\\
KScratch-7 & X\_Maximum=[172,227),Steel\_Plate\_Thickness=[-Inf,40.14) & 0.9436364 & 519 & 6.605454 & 0.1101677\\
KScratch-8 & Sum\_of\_Luminosity=[3.75e+05, Inf] & 0.9168111 & 529 & 6.417678 & 0.1122904\\
KScratch-9 & X\_Minimum=[36.9,44) & 0.9143426 & 459 & 6.400398 & 0.0974315\\
KScratch-10 & X\_Maximum=[172,227),TypeOfSteel=A400 & 0.9029463 & 521 & 6.320624 & 0.1105922\\
\addlinespace
Pastry-1 & X\_Minimum=[1.45e+03, Inf],Steel\_Plate\_Thickness=[80.07,99.97) & 1.0000000 & 85 & 7.000000 & 0.0180429\\
Pastry-2 & X\_Maximum=[1.47e+03, Inf],Steel\_Plate\_Thickness=[80.07,99.97) & 1.0000000 & 81 & 7.000000 & 0.0171938\\
Pastry-3 & X\_Minimum=[1.45e+03, Inf],Edges\_Y\_Index=[1, Inf],LogOfAreas=[2.11,2.59),TypeOfSteel=A400 & 1.0000000 & 61 & 7.000000 & 0.0129484\\
Pastry-4 & X\_Maximum=[1.47e+03, Inf],Edges\_Y\_Index=[1, Inf],LogOfAreas=[2.11,2.59),TypeOfSteel=A400 & 1.0000000 & 58 & 7.000000 & 0.0123116\\
Pastry-5 & Length\_of\_Conveyer=[1694.06,1705),Steel\_Plate\_Thickness=[80.07,99.97),SigmoidOfAreas=[0.445,0.978) & 1.0000000 & 55 & 7.000000 & 0.0116748\\
\addlinespace
Pastry-6 & Length\_of\_Conveyer=[1694.06,1705),Steel\_Plate\_Thickness=[80.07,99.97),TypeOfSteel=A400 & 0.9886364 & 87 & 6.920454 & 0.0184674\\
Pastry-7 & X\_Minimum=[1.45e+03, Inf],Length\_of\_Conveyer=[1694.06,1705) & 0.9863014 & 72 & 6.904110 & 0.0152834\\
Pastry-8 & X\_Maximum=[1.47e+03, Inf],Length\_of\_Conveyer=[1694.06,1705) & 0.9859155 & 70 & 6.901408 & 0.0148588\\
Pastry-9 & X\_Minimum=[1.45e+03, Inf],Pixels\_Areas=[131,472),Edges\_Y\_Index=[1, Inf],TypeOfSteel=A400 & 0.9850746 & 66 & 6.895522 & 0.0140098\\
Pastry-10 & X\_Maximum=[1.47e+03, Inf],Pixels\_Areas=[131,472),Edges\_Y\_Index=[1, Inf],TypeOfSteel=A400 & 0.9843750 & 63 & 6.890625 & 0.0133730\\
\addlinespace
Stains-1 & LogOfAreas=[0.656,1.38) & 0.9966555 & 596 & 6.976589 & 0.1265124\\
Stains-2 & Pixels\_Areas=[4,28.6) & 0.9919743 & 618 & 6.943820 & 0.1311823\\
Stains-3 & Sum\_of\_Luminosity=[764,2.91e+03) & 0.9896552 & 574 & 6.927586 & 0.1218425\\
Stains-4 & SigmoidOfAreas=[0.121,0.15) & 0.9892280 & 551 & 6.924596 & 0.1169603\\
Stains-5 & X\_Perimeter=[2.5,7.995) & 0.9791183 & 422 & 6.853828 & 0.0895776\\
\addlinespace
Stains-6 & Y\_Perimeter=[1.5,5.99) & 0.9676768 & 479 & 6.773737 & 0.1016769\\
Stains-7 & Log\_Y\_Index=[0.478,0.603) & 0.9537954 & 289 & 6.676568 & 0.0613458\\
Stains-8 & Steel\_Plate\_Thickness=[49.83,51.06) & 0.9353024 & 665 & 6.547117 & 0.1411590\\
ZScratch-1 & Length\_of\_Conveyer=[1354.01,1356),Steel\_Plate\_Thickness=[69.96,70.31) & 1.0000000 & 153 & 7.000000 & 0.0324772\\
ZScratch-2 & X\_Maximum=[14,164),Maximum\_of\_Luminosity=[103.027,118.024),Steel\_Plate\_Thickness=[69.96,70.31) & 0.9852941 & 134 & 6.897059 & 0.0284441\\
\addlinespace
ZScratch-3 & X\_Maximum=[14,164),Steel\_Plate\_Thickness=[69.96,70.31),Empty\_Index=[0.495,0.636) & 0.9677419 & 120 & 6.774193 & 0.0254723\\
ZScratch-4 & Length\_of\_Conveyer=[1356,1356.03),Steel\_Plate\_Thickness=[69.96,70.31) & 0.9673203 & 148 & 6.771242 & 0.0314158\\
ZScratch-5 & X\_Maximum=[14,164),Steel\_Plate\_Thickness=[69.96,70.31),Square\_Index=[0.5,0.681) & 0.9663866 & 115 & 6.764706 & 0.0244110\\
ZScratch-6 & Length\_of\_Conveyer=[1354.01,1356),TypeOfSteel=A300 & 0.9622642 & 153 & 6.735849 & 0.0324772\\
ZScratch-7 & Length\_of\_Conveyer=[1354,1354.01),Steel\_Plate\_Thickness=[69.96,70.31) & 0.9611650 & 99 & 6.728155 & 0.0210146\\
\addlinespace
ZScratch-8 & X\_Minimum=[111,295),Steel\_Plate\_Thickness=[69.96,70.31) & 0.9555556 & 215 & 6.688889 & 0.0456379\\
ZScratch-9 & X\_Maximum=[14,164),Minimum\_of\_Luminosity=[74,88.9),Steel\_Plate\_Thickness=[69.96,70.31) & 0.9496403 & 132 & 6.647482 & 0.0280195\\
ZScratch-10 & Steel\_Plate\_Thickness=[69.96,70.31),Empty\_Index=[0.495,0.636),TypeOfSteel=A300 & 0.9463415 & 194 & 6.624390 & 0.0411802\\
\bottomrule
\end{tabular}}
\end{table}

\textbf{Comments on Rules}\\
1. Length of RF-3 is too prohibitive.\\
2. Rules for K\_Scratch and Stains have highest counts and confidence, however at the same time they have smallest lengths. Thus making these rules very attractve insights for human use.

\hypertarget{mdl}{%
\section{Final Model Building}\label{mdl}}

Insights alongwith the extracted rules from RF and associations mining may be incorporated into a data frame for computing implementation. An example may be put here for explaining the rule-classifier ensamble as used in this work:\\
Suppose following 5 rules have been decided to be included in the ensamble:\\
1. IF \(V_{1}\) is \(A_{V1}\), \(V_{3}\) is \(C_{V3}\) THEN \(R_{1}\)\\
2. IF \(V_{2}\) is \(A_{V3}\), \(V_{3}\) is \(C_{V3}\) THEN \(R_{2}\)\\
3. IF \(V_{1}\) is \(B_{V1}\), \(V_{1}\) is \(A_{V1}\), \(V_{3}\) is \(A_{V3}\) THEN \(R_{3}\)\\
4. IF \(V_{2}\) is \(B_{V3}\), \(V_{3}\) is \(C_{V3}\) THEN \(R_{4}\)\\
5. IF \(V_{1}\) is \(C_{V1}\), \(V_{2}\) is \(B_{V2}\), \(V_{3}\) is \(B_{V3}\) THEN \(R_{5}\)

In present case \(A_{V1}\) etc are the semi-open intervals appearing in rules listings of Table \ref{tab:rulesListing}. Next, these may be re-written as the following table (a data frame):

\[
\begin{tabular}{cccc}
\hline
$\mathbf{V}_{1}$ & $\mathbf{V}_{2}$ & $\mathbf{V}_{3}$ & \textbf{Result} \\ \hline
$A_{V_{1}}$ & $-$ & $C_{V_{3}}$ & $R_{1}$ \\ 
$-$ & $A_{V_{2}}$ & $C_{V_{3}}$ & $R_{2}$ \\ 
$B_{V_{1}}$ & $A_{V_{2}}$ & $A_{V_{3}}$ & $R_{3}$ \\ 
$-$ & $B_{V_{3}}$ & $C_{V_{3}}$ & $R_{4}$ \\ 
$C_{V_{1}}$ & $B_{V_{2}}$ & $B_{V_{3}}$ & $R_{5}$ \\ \hline
\end{tabular}%
\]

\begin{figure}[h]
\includegraphics{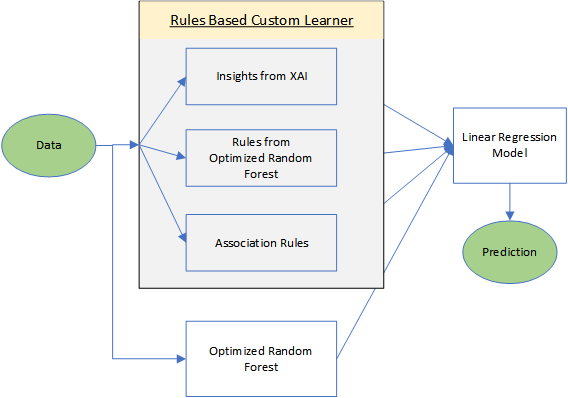} \caption{Schematic Diagram of High Precision Ensamble Classifier}\label{fig:highPrecisionClassifier}
\end{figure}

The Directed Acyclic Graph for development of the customized model based upon XAI insights combined with rules from RF and Association Rules Mining (ARM) is shown in Figure (\ref{fig:highPrecisionClassifier}):\\

Performance of the developed model is given as:\\

\begin{table}[H]

\caption{\label{tab:unnamed-chunk-5}Performance Comparision}
\centering
\begin{tabular}[t]{lrr}
\toprule
Model & Original & SMOTE\\
\midrule
Custom Learner & 0.9236 & 0.9418\\
Optimized RF & 0.7831 & 0.8923\\
\bottomrule
\end{tabular}
\end{table}

Same Confusion matrix for our developed classifier may be visualized as follows:\\

\begin{figure}[h]
\includegraphics{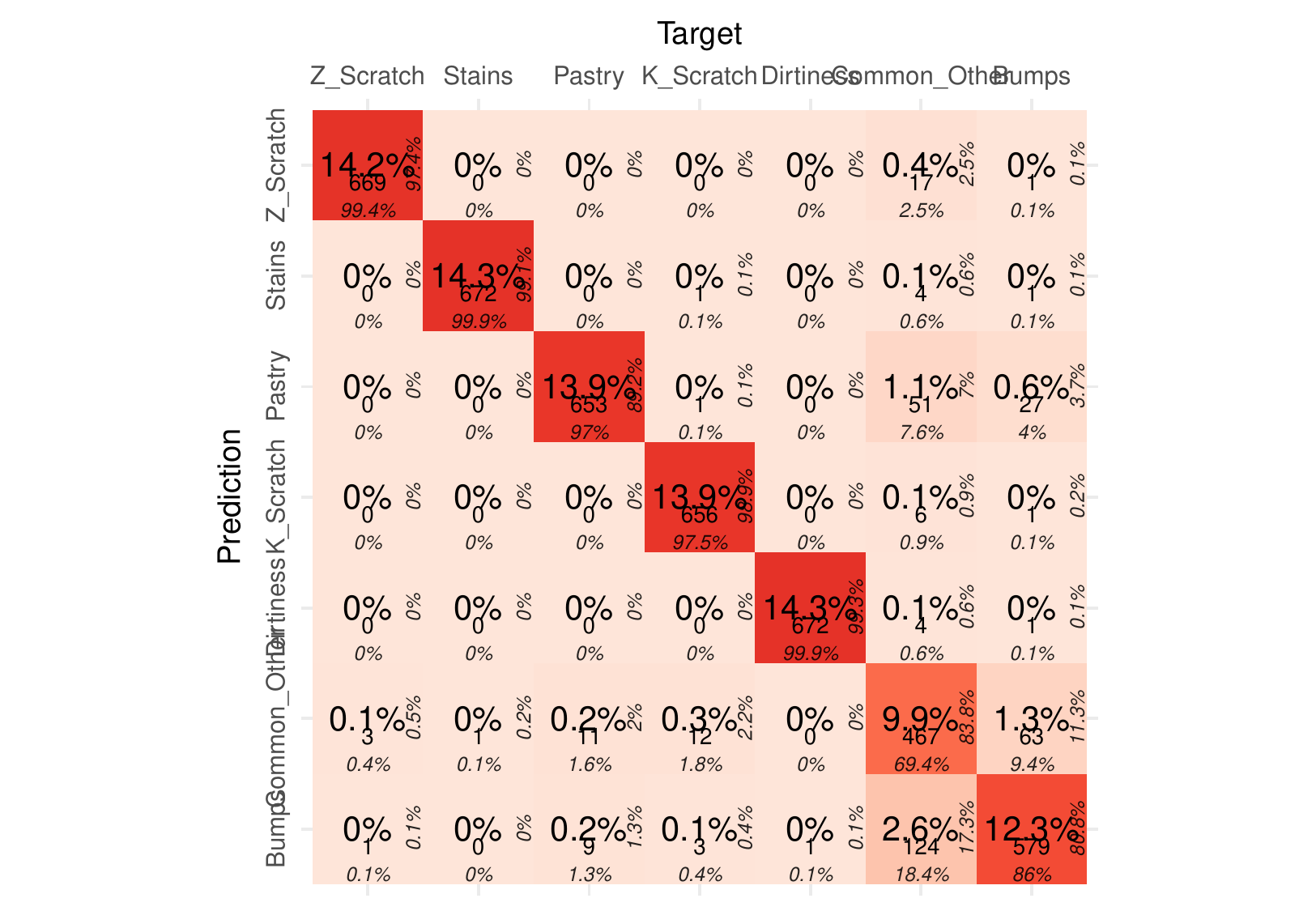} \caption{Confusion Matrix of the Developed Classifier}\label{fig:myconfmat}
\end{figure}

\begin{table}[H]

\caption{\label{tab:unnamed-chunk-7}Error for Custom Learner}
\centering
\begin{tabular}[t]{lr}
\toprule
Fault & Error\\
\midrule
Bumps & 6.03\\
Common\_Other & 5.21\\
Dirtiness & 4.40\\
K\_Scratch & 6.30\\
Pastry & 5.70\\
\addlinespace
Stains & 3.80\\
Z\_Scratch & 4.40\\
\bottomrule
\end{tabular}
\end{table}

\hypertarget{discuss}{%
\section{Discussion and Conclusions}\label{discuss}}

This work makes three contributions viz: insights, methodology and a high precision classifier. First of all it obtains valuable-for-humans insights using XAI, secondly it introduces a methodology to incorporate insights from three areas to build a high precision fault classifier. This high precision classifier is the third contribution.

Use of Synthetic Minority Oversampling Technique to balance the multi-class imbalance and the use of Medoids for taking full advantage of the local model agnostic XAI tools like Ceteris Peribus profiles and Breakdown plots are also important methodological innovations of this work.

The considerably improved performance of the classifier, in our view, is the result of few basic factors of this research. Idea of using various class balancing techniques for the problem of steel plate faults dataset was originally proposed by (Tian, Fu, and Wu 2015) in their conclusion of the paper as a future direction. However, this they proposed while ignoring all the 673 cases of Common\_Other type of faults.

In literature one finds a considerable number of sythetic data generation and class balancing techniques (Laopez et al. 2013).

A suitable choice of prototypical cases (Gurumoorthy et al. 2019) have been demonstrated to be useful for the purpose of explainability. Our present work makes a suitable choice of the representative, equivalently central and prototypical, cases as medoids. This choice of medoids allows for peeking into the global behaviour of the trained model while using a localized technique, for example, Ceteris Peribus.

Representing collection of rules as a data frame has been in practice for various practical use cases e.g.~illustrative example in (Deng 2019). Present work makes some changes for its own usage to cobble various If-Then rules obtained from XAI Insights, Random Forest and Association Mining.

Insights help to do Science (and Engineering, of course) `by' Data. The insights obtained in this work, at the very first place, are themselves directly useful for the production engineers. These insights provide for sharpening of the skill of humans to detect a fault and its relevant factors/variables.

Usually XAI is used to develop white-box models for human consumption. However in this work XAI has been used in a less trodden manner i.e.~getting insights from XAI and incorporating them back into rule-based ensemble development. These insights have been re-entered into the loop of model building which has clearly benefited the new model by improving its performance by almost 7\%.

This approach has also benefited from the wholistic view developed by Baniecki et. al.~(Baniecki and Biecek 2020). Techniques labeled as 1-4 were applied from Figure 2 (reproduced below) of Baniecki paper.

\begin{figure}[h]
\includegraphics{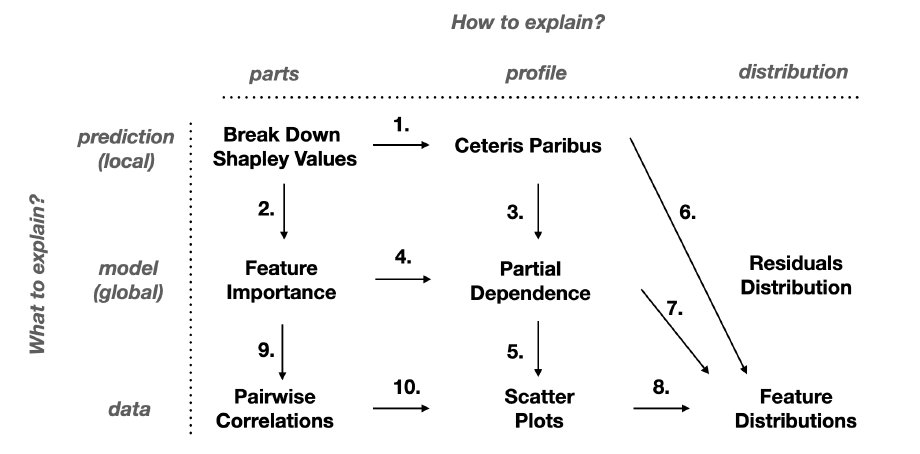} \caption{A Grammar of XAI tools taken from  (Baniecki and Biecek 2020): Present work has made use of paths 1, 2, 3, 4 and 9. Residual distributions are only applicable for learners with probablity outputs instead of actual responses.}\label{fig:unnamed-chunk-8}
\end{figure}

Certain areas remained unaddressed as well. There are other than SMOTE techinques available for oversampling and balancing of data (Laopez et al. 2013). These techniques should also be checked for even higher accuracy of the developed classifier. Another approach may be to develop a surrogate model for obtaining even more transparancy. Although in view of an Optimized Random Forest and its extracted rules this appears not too useful, but the quality measures of RF rules indicates a possible direction of improvement.

\hypertarget{references}{%
\section*{References}\label{references}}
\addcontentsline{toc}{section}{References}

\hypertarget{refs}{}
\leavevmode\hypertarget{ref-banieckiGrammarInteractiveExplanatory2020}{}%
Baniecki, Hubert, and Przemyslaw Biecek. 2020. ``The Grammar of Interactive Explanatory Model Analysis.'' \emph{arXiv:2005.00497 {[}Cs, Stat{]}}, May. \url{http://arxiv.org/abs/2005.00497}.

\leavevmode\hypertarget{ref-barredoarrietaExplainableArtificialIntelligence2020}{}%
Barredo Arrieta, Alejandro, Natalia Díaz-Rodríguez, Javier Del Ser, Adrien Bennetot, Siham Tabik, Alberto Barbado, Salvador Garcia, et al. 2020. ``Explainable Artificial Intelligence (XAI): Concepts, Taxonomies, Opportunities and Challenges Toward Responsible AI.'' \emph{Information Fusion} 58 (June): 82--115. \url{https://doi.org/10.1016/j.inffus.2019.12.012}.

\leavevmode\hypertarget{ref-chawla_smote_2002}{}%
Chawla, N. V., K. W. Bowyer, L. O. Hall, and W. P. Kegelmeyer. 2002. ``SMOTE: Synthetic Minority over-Sampling Technique.'' \emph{Journal of Artificial Intelligence Research} 16 (June): 321--57. \url{https://doi.org/10.1613/jair.953}.

\leavevmode\hypertarget{ref-christinab.azodiOpeningBlackBox2020}{}%
Christina B. Azodi, Jiliang Tang, and Shin-Han Shiu. 2020. ``Opening the Black Box: Interpretable Machine Learning for Geneticists.'' \url{https://doi.org/10.1016/j.tig.2020.03.005}.

\leavevmode\hypertarget{ref-consoliProducingLinkedData2017}{}%
Consoli, Sergio, Valentina Presutti, Diego Reforgiato Recupero, Andrea G. Nuzzolese, Silvio Peroni, Misael Mongiovi', and Aldo Gangemi. 2017. ``Producing Linked Data for Smart Cities: The Case of Catania.'' \emph{Big Data Research} 7 (March): 1--15. \url{https://doi.org/10.1016/j.bdr.2016.10.001}.

\leavevmode\hypertarget{ref-dengInterpretingTreeEnsembles2019}{}%
Deng, Houtao. 2019. ``Interpreting Tree Ensembles with inTrees.'' \emph{International Journal of Data Science and Analytics} 7 (4): 277--87. \url{https://doi.org/10.1007/s41060-018-0144-8}.

\leavevmode\hypertarget{ref-dengCBCAssociativeClassifier2014}{}%
Deng, Houtao, George Runger, Eugene Tuv, and Wade Bannister. 2014. ``CBC: An Associative Classifier with a Small Number of Rules.'' \emph{Decision Support Systems} 59 (March): 163--70. \url{https://doi.org/10.1016/j.dss.2013.11.004}.

\leavevmode\hypertarget{ref-desilvaIntelligentIndustrialInformatics2020}{}%
De Silva, Daswin, Seppo Sierla, Damminda Alahakoon, Evgeny Osipov, Xinghuo Yu, and Valeriy Vyatkin. 2020. ``Toward Intelligent Industrial Informatics: A Review of Current Developments and Future Directions of Artificial Intelligence in Industrial Applications.'' \emph{IEEE Industrial Electronics Magazine} 14 (2): 57--72. \url{https://doi.org/10.1109/MIE.2019.2952165}.

\leavevmode\hypertarget{ref-e.l.russellFaultDetectionIndustrial2000}{}%
E.L. Russell, L.H.Chiang, and R.D.Braatz. 2000. ``Fault Detection in Industrial Processes Using Canonical Variate Analysis and Dynamic Principal Component Analysis'' 51 (1): 81--93.

\leavevmode\hypertarget{ref-friedmanGreedyFunctionApproximation2001}{}%
Friedman, Jerome H. 2001. ``Greedy Function Approximation: A Gradient Boosting Machine.'' \emph{Annals of Statistics} 29 (5): 1189--1232. \url{https://doi.org/10.1214/aos/1013203451}.

\leavevmode\hypertarget{ref-RecognizingMaterialProperties2020}{}%
Gabriel Schwartz, and Ko Nishino. 2020. ``Recognizing Material Properties from Images,'' August. \url{https://doi.org/10.1109/TPAMI.2019.2907850}.

\leavevmode\hypertarget{ref-goldsteinPeekingBlackBox2015}{}%
Goldstein, Alex, Adam Kapelner, Justin Bleich, and Emil Pitkin. 2015. ``Peeking Inside the Black Box: Visualizing Statistical Learning with Plots of Individual Conditional Expectation.'' \emph{Journal of Computational and Graphical Statistics} 24 (1): 44--65. \url{https://doi.org/10.1080/10618600.2014.907095}.

\leavevmode\hypertarget{ref-grabotRuleMiningMaintenance2020}{}%
Grabot, Bernard. 2020. ``Rule Mining in Maintenance: Analysing Large Knowledge Bases.'' \emph{Computers \& Industrial Engineering} 139 (January): 105501. \url{https://doi.org/10.1016/j.cie.2018.11.011}.

\leavevmode\hypertarget{ref-gurumoorthyEfficientDataRepresentation2019}{}%
Gurumoorthy, Karthik S., Amit Dhurandhar, Guillermo Cecchi, and Charu Aggarwal. 2019. ``Efficient Data Representation by Selecting Prototypes with Importance Weights.'' \emph{arXiv:1707.01212 {[}Cs, Stat{]}}, August. \url{http://arxiv.org/abs/1707.01212}.

\leavevmode\hypertarget{ref-kangModelValidationFailure2020}{}%
Kang, Seokho. 2020. ``Model Validation Failure in Class Imbalance Problems.'' \emph{Expert Systems with Applications} 146 (May): 113190. \url{https://doi.org/10.1016/j.eswa.2020.113190}.

\leavevmode\hypertarget{ref-kazemiQualityControlClassification2018}{}%
Kazemi, Mohammad Ali Afshar, Sima Hajian, and Neda Kiani. 2018. ``Quality Control and Classification of Steel Plates Faults Using Data Mining.'' \emph{Applied Mathematics \& Information Sciences Letters} 6 (2): 59--67. \url{https://doi.org/10.18576/amisl/060202}.

\leavevmode\hypertarget{ref-kharalNeuralNetworksBased2012}{}%
Kharal, Athar, and Ayman Saleem. 2012. ``Neural Networks Based Airfoil Generation for a Given Cp Using BezierPARSEC Parameterization.'' \emph{Aerospace Science and Technology} 23 (1): 330--44. \url{https://doi.org/10.1016/j.ast.2011.08.010}.

\leavevmode\hypertarget{ref-lopezInsightClassificationImbalanced2013}{}%
Laopez, Victoria, Alberto Fernandez, Salvador Garcia, Vasile Palade, and Francisco Herrera. 2013. ``An Insight into Classification with Imbalanced Data: Empirical Results and Current Trends on Using Data Intrinsic Characteristics.'' \emph{Information Sciences} 250 (November): 113--41. \url{https://doi.org/10.1016/j.ins.2013.07.007}.

\leavevmode\hypertarget{ref-nkonyanaPerformanceEvaluationData2019}{}%
Nkonyana, Thembinkosi, Yanxia Sun, Bhekisipho Twala, and Eustace Dogo. 2019. ``Performance Evaluation of Data Mining Techniques in Steel Manufacturing Industry.'' \emph{Procedia Manufacturing}, The 2nd International Conference on Sustainable Materials Processing and Manufacturing, SMPM 2019, 8-10 March 2019, Sun City, South Africa, 35 (January): 623--28. \url{https://doi.org/10.1016/j.promfg.2019.06.004}.

\leavevmode\hypertarget{ref-panLearningImbalancedDatasets2020}{}%
Pan, Tingting, Junhong Zhao, Wei Wu, and Jie Yang. 2020. ``Learning Imbalanced Datasets Based on SMOTE and Gaussian Distribution.'' \emph{Information Sciences} 512 (February): 1214--33. \url{https://doi.org/10.1016/j.ins.2019.10.048}.

\leavevmode\hypertarget{ref-r.berkStatisticalProceduresForecasting2013}{}%
R. Berk, and J. Bleich. 2013. ``Statistical Procedures for Forecasting Criminal Behavior: A Comparative Assessment.'' \emph{Criminology and Public Policy} 12: 513--44.

\leavevmode\hypertarget{ref-roscherExplainableMachineLearning2020}{}%
Roscher, Ribana, Bastian Bohn, Marco F. Duarte, and Jochen Garcke. 2020. ``Explainable Machine Learning for Scientific Insights and Discoveries.'' \emph{IEEE Access} 8: 42200--42216. \url{https://doi.org/10.1109/ACCESS.2020.2976199}.

\leavevmode\hypertarget{ref-santiagoegeagomezExploratoryStudyClass2019}{}%
Santiago Egea Gomez, Luis Hernandez-Callejo, Belen Carro Martinez, and Antonio J. Sanchez-Esguevillas. 2019. ``Exploratory Study on Class Imbalance and Solutions for Network Traffic Classification.'' \emph{Neurocomputing}, Learning in the Presence of Class Imbalance and Concept Drift, 343 (May): 100--119. \url{https://doi.org/10.1016/j.neucom.2018.07.091}.

\leavevmode\hypertarget{ref-semeionSteelPlatesFaults}{}%
Semeion. n.d. ``Steel Plates Faults Data Set.'' Repository. \emph{UCI Machine Learning Repository}. https://archive.ics.uci.edu/ml/datasets/Steel+Plates+Faults.

\leavevmode\hypertarget{ref-soylemezogluMahalanobisTaguchiSystemMultiSensor2011}{}%
Soylemezoglu, Ahmet, Sarangapani Jagannathan, and Can Saygin. 2011. ``Mahalanobis-Taguchi System as a Multi-Sensor Based Decision Making Prognostics Tool for Centrifugal Pump Failures.'' \emph{IEEE Transactions on Reliability} 60 (4): 864--78. \url{https://doi.org/10.1109/TR.2011.2170255}.

\leavevmode\hypertarget{ref-tian_steel_2015}{}%
Tian, Yang, Mengyu Fu, and Fang Wu. 2015. ``Steel Plates Fault Diagnosis on the Basis of Support Vector Machines.'' \emph{Neurocomputing} 151 (March): 296--303. \url{https://doi.org/10.1016/j.neucom.2014.09.036}.

\leavevmode\hypertarget{ref-yuImprovedKmedoidsAlgorithm2018}{}%
Yu, Donghua, Guojun Liu, Maozu Guo, and Xiaoyan Liu. 2018. ``An Improved K-Medoids Algorithm Based on Step Increasing and Optimizing Medoids.'' \emph{Expert Systems with Applications} 92 (February): 464--73. \url{https://doi.org/10.1016/j.eswa.2017.09.052}.

\end{document}